\documentclass[lettersize,journal]{IEEEtran}
\usepackage{amssymb}
\usepackage{amsthm}
\usepackage[utf8]{inputenc}
\usepackage{amsmath}
\usepackage{graphicx} 
\usepackage{multirow}
\usepackage{graphics}
\usepackage{graphicx}
\usepackage{booktabs}
\usepackage{setspace}
\usepackage{array}
\usepackage{booktabs}
\usepackage{cite}
\usepackage[colorlinks]{hyperref}
\usepackage[table,xcdraw]{xcolor}

\usepackage{makecell}
\usepackage{tikz,xcolor,hyperref}
\usepackage[caption=false,font=normalsize,labelfont=sf,textfont=sf]{subfig}
\usepackage{bm}

\usepackage{calc}
\usepackage{pgf}
\usepackage[switch]{lineno}
\usepackage{algorithm}
\usepackage{algpseudocode}
\usepackage[most]{tcolorbox}
\usepackage[framemethod=TikZ]{mdframed}
\tcbuselibrary{listings, breakable}

\definecolor{LightRed}{rgb}{1,0.92,0.92}
\definecolor{LightOrange}{rgb}{1,0.95,0.88}
\definecolor{LightYellow}{rgb}{1.0,1.0,0.84}
\definecolor{LightGreen}{rgb}{0.9,1.0,0.88}
\definecolor{LightCyan}{rgb}{0.9,1,1}
\definecolor{LightBlue}{rgb}{0.9,0.94,1}
\definecolor{lime}{HTML}{A6CE39}
\usepackage{tabularx}

\DeclareRobustCommand{\orcidicon}{%
	\begin{tikzpicture}
		\draw[lime, fill=lime] (0,0) 
		circle [radius=0.16] 
		node[white] {{\fontfamily{qag}\selectfont \tiny ID}};    \draw[white, fill=white] (-0.0625,0.095) 
		circle [radius=0.007];    \end{tikzpicture}
	\hspace{-2mm}}
\foreach \x in {A, ..., Z}{%
	\expandafter\xdef\csname orcid\x\endcsname{\noexpand\href{https://orcid.org/\csname orcidauthor\x\endcsname}{\noexpand\orcidicon}}
}

\newtcolorbox{ServerBlock}{
	colback=blue!5, colframe=blue!40,
	sharp corners, boxrule=0.3pt,
	left=0pt, right=0pt, top=1pt, bottom=1pt,
	boxsep=0pt, arc=2pt
}

\newtcolorbox{ClientBlock}{
	colback=green!5, colframe=green!40,
	sharp corners, boxrule=0.3pt,
	left=0pt, right=0pt, top=1pt, bottom=1pt,
	boxsep=0pt, arc=2pt
}
\begin{document}

\title{Token-Level Prompt Mixture with Parameter-Free Routing for Federated Domain Generalization}

\author{
	Shuai Gong\orcidB{}, Chaoran Cui\textsuperscript{*}\orcidC{}, Xiaolin Dong\orcidD{}, Xiushan Nie\orcidE{}, Senior Member, IEEE, Lei Zhu\orcidF{}, Senior Member, IEEE, Xiaojun Chang\orcidG{}, Senior Member, IEEE
	\thanks{This work was supported by the Shandong Provincial
		Natural Science Foundation under Grant ZR2020KF015, and by the Taishan
		Scholar Program of Shandong Province under Grant tsqn202211199 and Grant
		tstp20221137.
		
		S. Gong, C. Cui, and X. Dong  are with the
		School of Computing and Artificial Intelligence, Shandong University of
		Finance and Economics, Jinan 250014, China (e-mail: gsh8210@163.com; crcui@sdufe.edu.cn;
		d\_lin1998@163.com).
		
		X. Nie is with the School of Computer Science and Technology,
		Shandong Jianzhu University, Jinan 250101, China (e-mail: niexsh@hotmail.com).
		
		L. Zhu is with the College of Electronics and Information Engineering,
		Tongji University, Shanghai 201804, China (e-mail: leizhu0608@gmail.com).
		
		X. Chang is with the School of Information Science and Technology, University of Science and Technology of China, Anhui 230026, China.   (e-mail: cxj273@gmail.com). 
	}	
	
	\thanks{\textsuperscript{*}Corresponding author.}
}

\markboth{Journal of \LaTeX\ Class Files,~Vol.~14, No.~8, August~2021}%
{Shell \MakeLowercase{\textit{et al.}}: A Sample Article Using IEEEtran.cls for IEEE Journals}


\maketitle
\begin{abstract}
Federated domain generalization (FedDG) aims to learn a globally generalizable model from decentralized clients with heterogeneous data while preserving privacy.  
Recent studies have introduced prompt learning to adapt vision-language models (VLMs) in FedDG by learning a single global prompt. 
However, such a one-prompt-fits-all learning paradigm typically leads to performance degradation on personalized samples.
Although the mixture of experts (MoE) offers a promising solution for specialization, existing MoE-based methods suffer from coarse image-level expert assignment and high communication costs from parameterized routers.
To address these limitations, we propose TRIP, a Token-level pRompt mIxture with Parameter-free routing framework for FedDG, which treats multiple prompts as distinct experts. 
Unlike existing image-level routing designs, TRIP assigns different tokens within an image to specific experts.  
To ensure communication efficiency, TRIP incorporates a parameter-free routing mechanism based on token clustering and optimal transport. 
The instance-specific prompt is then synthesized by aggregating experts, weighted by the number of tokens assigned to each.
Additionally, TRIP develops an unbiased learning strategy for prompt experts, leveraging the VLM’s zero-shot generalization capability. 
Extensive experiments across four benchmarks demonstrate that TRIP achieves optimal generalization results, with communicating only 1K parameters per round.
Our code is available at \url{https://github.com/GongShuai8210/TRIP}.
\end{abstract}

\begin{IEEEkeywords}
Federated Domain Generalization, Prompt Learning, Mixture of Experts, Optimal Transport.
\end{IEEEkeywords}

\section{Introduction}
\IEEEPARstart {T}{he} exponential growth of data from diverse and decentralized sources has significantly accelerated advancements in machine learning.
However, traditional machine learning paradigms typically operate in a centralized manner, which requires all decentralized data to be processed on a central server, raising serious privacy concerns.
Recently, federated learning (FL)~\cite{mcmahan2017communication,li2021survey} has emerged as a prevalent paradigm to address this issue effectively. 
Unlike its centralized counterparts, FL allows distributed clients to train  local models on their respective data. 
Subsequently, these models are aggregated at a central server to update a global model.

Despite the effectiveness of FL, it often assumes that client data are independently and identically distributed (IID).
In real-world scenarios, however, clients frequently exhibit non-IID data distributions, naturally forming multiple source domains.
When deploying federated models trained on these source domains to a novel, unseen target domain, the distribution shift poses a significant generalization challenge.  
This issue is extensively studied under the umbrella of domain generalization (DG)~\cite{zhou2024mixstyle,dayal2024madg},
which seeks to train robust models across diverse source domains to ensure generalizability to novel target domains.
However,  applying  DG techniques directly in FL settings is non-trivial, as they typically require centralized access to all source client data---violating the core principles of FL.  
This limitation promotes the emergence of federated domain generalization  (FedDG)~\cite{liu2021feddg}, which enables collaborative learning of a generalizable model from decentralized clients within the FL paradigm.

Recently, vision-language models (VLMs)~\cite{radford2021learning,li2022blip} have attracted significant attention due to their strong generalization capabilities.
To better adapt the large-scale VLMs to downstream tasks, a parameter-efficient technique called prompt learning~\cite{zhou2022learning,khattak2023maple} has emerged. 
Instead of adjusting all of the model's parameters, prompt learning only optimizes a small set of continuous embeddings inserted into the inputs.
Due to its ability to reduce training costs while further improving the generalization of VLMs, prompt learning is widely studied for FedDG tasks~\cite{bai2024diprompt,gong2024federated}.
\begin{figure*}[t]
	\centering
	\includegraphics[width=1\linewidth]{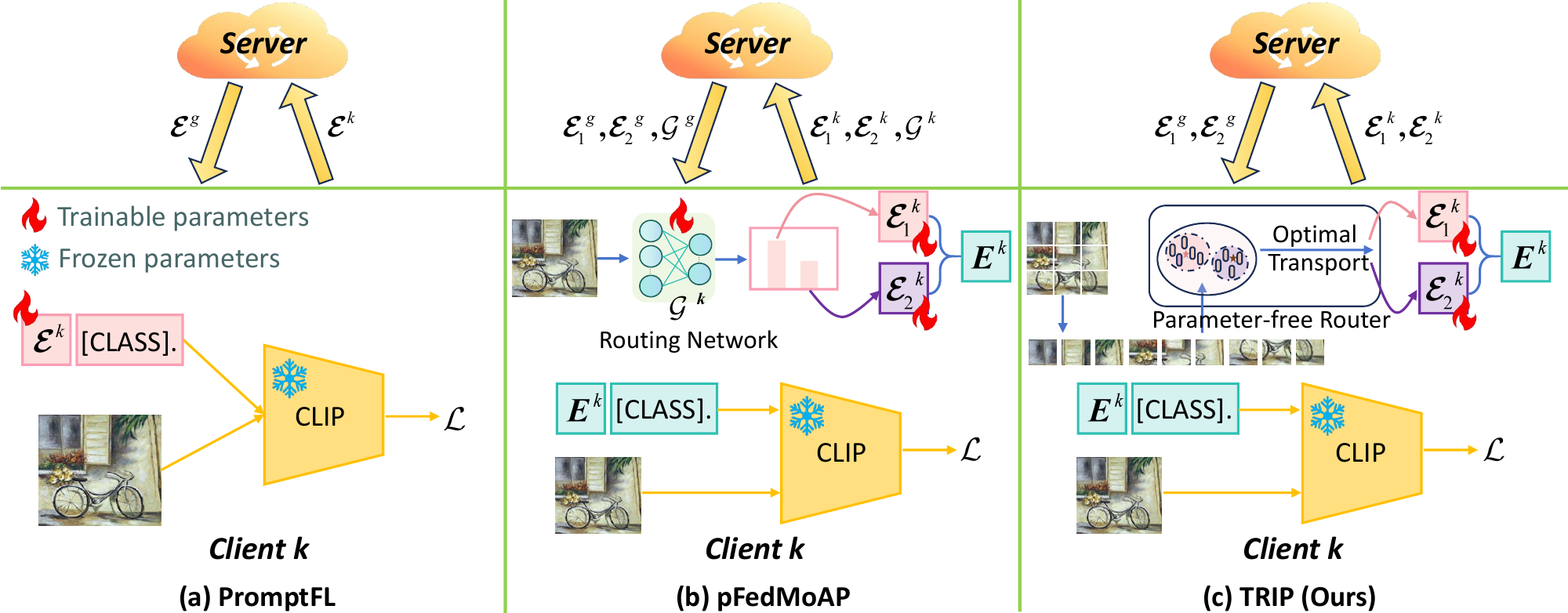}
	\captionsetup{skip=-1pt}
	\caption{Comparison of our method, TRIP, with prior prompt-based methods.
		(a) PromptFL~\cite{guo2023promptfl} aims to learn a shared global prompt applied to all samples, which limits its generalization ability.
		(b) pFedMoAP~\cite{luo2024mixture} trains a parameterized routing network to assign personalized prompts for each sample, but this introduces additional parameters that must be communicated between clients and the server.
		(c) Our method constructs a parameter-free router to assign image tokens to prompt experts, significantly reducing communication overhead while improving generalization performance.}
	\label{fig:head}
\end{figure*}

Prior prompt learning-based FedDG methods typically focus on learning a shared global prompt~\cite{guo2023promptfl,gong2024federated}, with the goal of guiding VLMs to adapt to all client data.
However, due to data heterogeneity across domains, such a one-prompt-fits-all learning paradigm may struggle to effectively capture personalized characteristics of the data.
One promising approach to address this challenge is to enable instance-specific adaptation~\cite{feng2023towards}, in which each image is processed using a personalized prompt, thus circumventing the constraints of a single global prompt.
Mixture of experts (MoE)~\cite{jacobs1991adaptive,dai2024deepseekmoe}
provides a natural framework for such instance-specific modeling, employing multiple specialized sub-models (referred to as ``experts") that collaborate to make predictions. Guided by a routing mechanism, MoE selectively activates and aggregates the most relevant experts for each input instance.

A few recent studies~\cite{luo2024mixture,wu2024mixture} have begun to explore prompt learning  with MoE. These methods learn a set of prompts, where each prompt acts as an expert.
Despite effectiveness, there are
two major limitations remain: 1) Traditional MoE frameworks typically route entire input instances to specific experts based on global image features. 
This coarse-grained routing, however, may not fully capture the fine-grained diversity within an image, where different regions of an image may require specialized experts to process; and
2) The routing mechanism relies on training a parameterized routing network to assign experts. Nevertheless, the trainable network inherently introduces additional parameters that require communication between the server and clients, thereby increasing communication overhead in FL.

To address the aforementioned limitations, we introduce a novel Token-level pRompt mIxture with Parameter-free routing framework for FedDG, dubbed as TRIP. 
Fig.~\ref{fig:head} compares previous prompt-based FedDG methods with TRIP. 
In contrast to existing image-level routing designs~\cite{luo2024mixture}, TRIP operates at the token level, assigning different tokens within an image to distinct experts. 
This token-level approach allows distinct local image regions to be processed by specialized experts, thereby facilitating the capture of fine-grained visual patterns.

To ensure communication efficiency, TRIP incorporates a parameter-free routing mechanism based on token clustering and optimal transport (OT)~\cite{monge1781memoire}. 
First, TRIP performs capacity-aware token clustering to ensure balanced cluster workloads, such that each cluster's size (i.e., the number of tokens it contains) adheres to a predefined capacity. 
Second, the generated token clusters are assigned to prompt experts via OT. 
However, directly applying OT between clusters and experts presents a stability challenge: since prompt experts are continuously optimized during training, the OT cost matrix fluctuates. 
This fluctuation can lead to inconsistent assignments, where similar clusters might be assigned to different experts across iterations. 
To overcome this challenge, TRIP introduces a set of static, non-learnable ``keys'', with each key uniquely corresponding to a prompt expert. 
The core idea is to perform the OT assignment between the token clusters and these stable keys, rather than directly with the experts. 
Since the keys remain fixed, this ensures that semantically similar clusters are stably assigned to the same keys (and thus, to the same experts). 
Finally, image tokens are routed to experts based on the key assignments of their respective clusters. 
The instance-specific prompt for an image is then synthesized by combining all prompt experts, weighted by the proportion of that image's tokens routed to each expert.

Previous prompt-based studies in FL~\cite{guo2023promptfl,bai2024diprompt} primarily optimize prompts independently on each local client, resulting in local prompts biased towards each client’s
private data patterns. 
Such bias has been demonstrated to compromise generalization to novel domains~\cite{khattak2023self}.
To tackle this challenge, TRIP introduces a debiasing strategy that balances the specialization and generalization of local prompt experts.
Leveraging the strong generalization capabilities of VLMs, TRIP takes the VLM's zero-shot prediction distribution as an anchor, aligning predictions generated by local prompt experts with this distribution.
By doing so,  TRIP facilitates indirect synchronization of local prompt experts across clients, thereby mitigating bias toward local client data.
Once the optimization of prompt experts is completed, each client uploads its prompt experts to the server for global aggregation using the FedAvg algorithm~\cite{mcmahan2017communication}. The globally aggregated prompt experts are subsequently distributed back to the clients for the next training iteration.

Empirical evaluations across four standard benchmarks demonstrate that TRIP substantially outperforms conventional FL, DG, and FedDG methods, along with existing prompt learning-based methods.  Ablation studies confirm that TRIP’s token-level routing enables finer-grained expert specialization, leading to stronger generalization than image-level routing.  Moreover, visualizations provide qualitative evidence that different prompt experts effectively capture region-specific image features.

In a nutshell, our main contributions are as follows:
\begin{itemize}
	\item We propose TRIP, a token-level prompt mixture framework for FedDG, which incorporates a token-level routing mechanism, enabling prompt experts to capture finer-grained image information.

	\item  In contrast to conventional learnable routers, the routing mechanism in TRIP is  parameter-free, significantly improving communication efficiency.
	
	\item We propose a debiasing strategy that balances expert specialization and generalization by leveraging the zero-shot capabilities of VLMs.
	
 	\item With only 1K communicated parameters per round, TRIP achieves state-of-the-art performance on four benchmark datasets.
\end{itemize}

The remainder of the paper is structured as follows:
Section~\ref{sec:related_work} provides a review of related work.
Section~\ref{sec:preliminary} introduces the preliminary concepts essential to our study.
Section~\ref{sec:method} presents the proposed TRIP method in detail.
Section~\ref{sec:experiments} discusses the experimental results and analysis. Finally, Section~\ref{sec:conclusions} concludes the paper.

\section{Related Work}\label{sec:related_work}
\subsection{Federated Learning}
Federated learning (FL) has emerged as a popular paradigm in distributed machine learning, enabling multiple clients to collaboratively train a global model while preserving data privacy. 
As a pioneering work in FL, FedAvg~\cite{mcmahan2017communication} constructs a global model by aggregating locally trained model parameters, where aggregation weights are proportional to client dataset sizes.
However, its assumption of uniform client contributions poses a significant challenge when dealing with non-IID data
across clients. 
To address this issue, FedProx~\cite{li2020federated} introduces a proximal regularization term in the local optimization objective  to align client updates with the global model. 
Similarly, MOON~\cite{malaviya2023fedfame} and SCAFFOLD~\cite{karimireddy2020scaffold}  utilize the global model as an anchor to prevent local models from diverging too far. 
Another line of research focuses on refining aggregation strategies to tackle data heterogeneity.
FedMA~\cite{wang2020federated} reformulates aggregation as a layer-wise weight-matching problem rather than simple weighted averaging. 
In contrast, FedNova~\cite{wang2020tackling} normalizes local updates based on client-specific training steps to ensure fair aggregation.

The aforementioned methods primarily address data heterogeneity stemming from label distribution skew. 
However, clients may also exhibit feature distribution shifts due to variations in spatiotemporal contexts.  
FedBN~\cite{li2021fedbn} personalizes batch normalization layers, allowing local models to better adapt to their unique feature distributions. 
FedHEAL~\cite{chen2024fair} incorporates domain diversity during aggregation to reduce the global model's bias toward domain-specific data patterns.
Despite their effectiveness, existing FL methods primarily focus on adapting the global model to previously seen local clients, neglecting scenarios where the global model must generalize to unseen clients with significantly different data distributions.

\subsection{Federated Domain Generalization}
Domain generalization (DG) aims to learn a model from multiple source domains that is capable of generalizing to unseen target domains. A range of classical domain alignment approaches aim to learn invariant features by minimizing discrepancy across source domains, utilizing techniques like maximum mean discrepancy~\cite{li2018domain}, contrastive learning~\cite{motiian2017unified}, and adversarial learning~\cite{dayal2024madg}. Due to the inherent difficulty of domain alignment, ensemble learning-based approaches have emerged as an alternative. These approaches ensemble learned domain-specific modules, such as classifiers~\cite{zhoudomain} or batch normalization layers~\cite{seo2020learning}, to improve generalization. 
Additionally, meta-learning approaches \cite{li2018learning,dou2019domain} have gained considerable attention, seeking to learn robust semantic features through learning-to-learn paradigms. 
As these methods rely on a centralized learning framework that requires unrestricted access to all source domain data, the data privacy is compromised.  
This limitation has motivated the emergence of federated domain generalization (FedDG).

FedDG addresses data privacy concerns in DG by leveraging the FL paradigm, enabling model training across decentralized domains without the need for raw data sharing. 
A common idea in FedDG is to promote knowledge sharing between clients while maintaining privacy protection. 
Recent works have explored sharing domain-specific information among clients, such as image style information~\cite{chen2023federated,park2024stablefdg}, amplitude spectra~\cite{liu2021feddg}, and class prototypes~\cite{huang2023rethinking}.
Additionally, the method GA~\cite{zhang2023federated} also explores reducing global model bias toward specific clients to achieve fair federated aggregation. GA adjusts the aggregation weights based on the variance in clients' generalization gaps.
These methods typically require transmitting the entire model's parameters between clients and the server, resulting in heavy communication overhead. 
In contrast, our method only communicates prompt parameters, significantly reducing communication costs.
\subsection{Prompt Learning in VLMs}
Prompt learning introduces a small set of learnable prompt parameters as additional inputs while maintaining the model's weights frozen, offering an efficient way for adapting large-scale models to downstream tasks.
As a pioneering work, CoOp~\cite{zhou2022learning} introduces continuous learnable prompt vectors in the text encoder, demonstrating substantial  performance improvements across downstream image recognition tasks. 
CoCoOp~\cite{zhou2022conditional} further incorporates image information as an additional conditioning signal to guide the generation of text prompts,  significantly  improving generalization to novel categories.
Instead of optimizing prompts in the text side only, MaPLe~\cite{khattak2023maple}  introduces learnable visual prompts to enhance the alignment between vision and language representations.
Recent work has explored learning multiple prompts with OT~\cite{monge1781memoire}, aiming to capture the diverse characteristics of categories~\cite{chenplot} or achieve fine-grained alignment of multi-modal prompts~\cite{wang2024tuning}.

Prompt learning has also drawn attention within the FL community, as it naturally meets the communication efficiency demands of FL. 
PromptFL~\cite{guo2023promptfl} extends CoOp to the FL setting, where each client learns local prompts using its own data, followed by the aggregation of these prompts into global prompts using the FedAvg algorithm. 
FedOTP~\cite{li2024global} further employs OT to align local prompts with global prompts, effectively addressing the challenge of data heterogeneity.
These methods typically employ unified global prompts for each sample, which leads to a significant performance degradation when facing unknown, different data distributions. 
Our method conceptualizes prompts as experts in a MoE framework, employing a router to enable instance-specific prompt synthesis.

\subsection{Mixture of Experts}
The core concept of mixture of experts (MoE)  is to divide the model into multiple ``expert" submodules, each of which can be viewed as a portion of the neural network specializing in specific tasks or particular types of inputs~\cite{jacobs1991adaptive}. 
In this context, MoE layers are constructed with a collection of expert networks, along with a router that dynamically selects the appropriate experts  based on the given inputs. 
Among the diverse MoE variants, feed-forward networks (FFNs)  remain the dominant choice as experts, as seen in models such as Switch Transformer~\cite{fedus2022switch} and DeepSeekMoE~\cite{dai2024deepseekmoe}.
Furthermore, based on the router design, existing MoE research can be broadly categorized into two types: the dense MoE~\cite{wumixture,shahbaba2009nonlinear}, which activates all experts, and the sparse MoE~\cite{shazeer2017outrageously,fedus2022switch}, which selectively activates only the most relevant ones. 

MoE has also been integrated into FL to personalize models for better adaptation to local client data. The core idea is that clients utilize their local data to train individual routers and multiple experts, which are subsequently sent back to the server for aggregation, as demonstrated in FedJETs~\cite{dun2023fedjets} and FedMix~\cite{reisser2021federated}. 
More recently, pFedMoAP~\cite{luo2024mixture} has explored the integration of prompt learning with the MoE framework, which treats prompts as experts and employs an attention-based router to dynamically generate personalized prompts for individual images.
Despite being effective, the method operates at the image level, potentially limiting the experts' ability to capture fine-grained regional features of input images. Moreover, it relies on a parameterized routing network that requires frequent communication between the server and clients, introducing significant communication overhead. 
Our method introduces a parameter-free router that operates at the token level, effectively addressing these limitations while maintaining model performance.
\section{Preliminaries}\label{sec:preliminary}
In this section, we first  provide a concise overview of CLIP, and then present a formal definition of FedDG.

\subsection{Revisiting CLIP}
As a classic VLM, CLIP~\cite{radford2019language} is famous for its remarkable zero-shot inference performance.
CLIP consists of a text encoder and a vision encoder.
Its text encoder processes textual descriptions containing image class names, formatted as ``a photo of a [CLASS].'', to generate a text feature $\bm{w}$. Meanwhile, the vision encoder extracts the visual feature $\bm{f}$ from an input image.
To enable CLIP for image classification tasks, the text encoder generates a set of text features  $\left\{\bm{w}_1, \bm{w}_2, \ldots, \bm{w}_C\right\}$ corresponding to $C$ image categories. For an image with visual feature $\bm{f}$, its probability of belonging to category $c$ can be calculated by:
\begin{equation}\label{eq:clip}
	p\left( {y = c|\bm{x}} \right) = \frac{{\exp \left( {\cos \left( {{\bm{w}_c},\bm{f}} \right)/\tau } \right)}}{{\sum\nolimits_{j = 1}^C {\exp \left( {\cos \left( {{\bm{w}_j},\bm{f}} \right)/\tau } \right)} }},
\end{equation}
where $\tau$ is a learnable temperature parameter, and $\cos(\cdot, \cdot)$ represents the cosine similarity.
\begin{figure*}[t]
	\centering
	\includegraphics[width=0.95\linewidth]{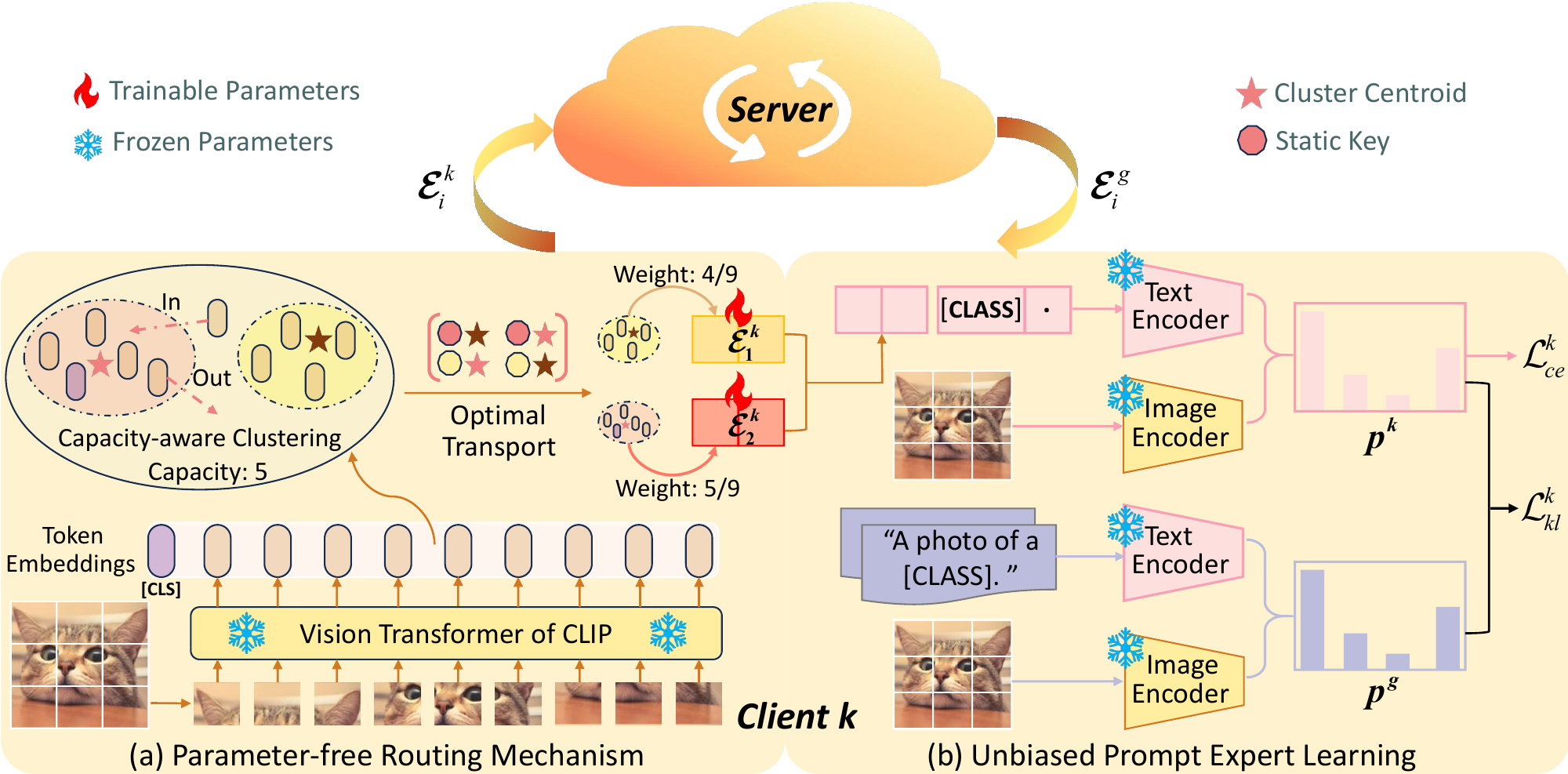}
	\captionsetup{skip=-1pt}
	\caption{Conceptual framework of TRIP for FedDG. TRIP consists of two essential components: (a) a parameter-free routing mechanism and (b) unbiased prompt expert learning.
		In (a), an image is first divided into patches and encoded into token embeddings. These embeddings are then grouped into clusters based on  capacity-aware clustering. Next, clusters are assigned to experts using OT by minimizing the total cost in a cost matrix constructed from the distances between cluster centroids and static keys.
		In (b), the local prediction distribution generated by local prompt experts are aligned with CLIP’s zero-shot inference distribution to mitigate potential bias toward localized data patterns.}
	\label{fig:framework}
\end{figure*}
\subsection{Problem Formulation for FedDG}
In accordance with standard practice in FedDG, we focus on the $C$-way image classification task, where the source data is distributed across $K$ clients.
Formally, let $(\mathcal{X}, \mathcal{Y})$ denote the joint image and label space.
The collection of source datasets is $\mathcal{S}=\{\mathcal{S}^k\}_{k=1}^{K}$, where each client $k$ holds a local dataset $\mathcal{S}^k=\{(\bm{x}^{k}_i,y^{k}_i) \in (\mathcal{X}, \mathcal{Y})\}_{i=1}^{|\mathcal{S}_k|}$ containing $|\mathcal{S}_k|$ image-label pairs. Due to factors like independent data collection, the data distribution $\mathbb{P}_{\mathcal{S}^k}$ associated with each client $k$ is generally distinct from others. 
The FedDG objective is to collaboratively learn a single global model $f:\mathcal{X} \rightarrow \mathcal{Y}$ from distributed sources $\{\mathcal{S}^k\}_{k=1}^{K}$ that generalizes effectively to unseen target domains, whose distributions $\{\mathbb{P}_{\mathcal{T}}\}$ may differ significantly from the source distributions $\{\mathbb{P}_{\mathcal{S}^k}\}_{k=1}^{K}$.
\section{Method}\label{sec:method}
Our proposed method, TRIP, introduces a parameter-free token-level MoE framework based on the capacity-aware clustering and OT.
In this framework, image tokens are first grouped into multiple clusters, where each cluster is constrained by a predefined capacity. These clusters are then assigned to prompt experts via OT.
Additionally, TRIP incorporates a debiasing strategy that calibrates the local prediction distributions by aligning them with CLIP's zero-shot prediction distribution.
The comprehensive overview of TRIP is presented in Fig.~\ref{fig:framework}.
\subsection{Capacity-Aware Clustering for Tokens}
Intuitively, semantically similar tokens should be routed to the same expert, as this allows each expert to specialize in processing specific semantic patterns. 
Building upon this insight, TRIP employs the \textit{k}\text{-}means algorithm to cluster the tokens derived from input images.
Specifically, for a given image $\bm{x}_i^k$ from client $k$, it is first processed by the CLIP vision encoder to generate a sequence of token embeddings, represented as the set $\left\{ \bm{x}_{i0}^k, \bm{x}_{i1}^k, \dots, \bm{x}_{iN}^k \right\}$. 
Each embedding $\bm{x}_{ij}^k \in \mathbb{R}^D$ is a $D$-dimensional vector. The first element, $\bm{x}_{i0}^k$, corresponds to the embedding of the special \texttt{[CLS]} token, while the remaining $N$ elements, $\left\{ \bm{x}_{ij}^k \right\}_{j=1}^N$, denote the embeddings of the $N$ image patches extracted from $\bm{x}_i^k$.
Our goal is to group these $N+1$ token embeddings into $M$ clusters by minimizing the standard \textit{k}\text{-}means objective function:
\begin{equation}\label{eq:k-means}
	\quad \sum_{m=1}^{M} \sum_{\bm{x}_{ij} \in \mathcal{C}_m^k} \| \bm{x}_{ij}^k - \bm{\mu}_m^k \|^2,
\end{equation}
where 
\( \mathcal{C}_m^k \) represents the set of tokens assigned to the $m$-th cluster at client $k$, and
\( \bm{\mu}_m^k \) is the centroid of the \( m \)-th cluster computed as the mean embedding of all tokens  in \( \mathcal{C}_m^k \).

After the clustering stage, TRIP assigns all tokens within each cluster to a dedicated expert.
Consequently, each expert’s token workload directly corresponds to the capacity of its assigned cluster.
Prior work~\cite{fedus2022switch} has shown that balanced expert workloads are vital for effective expert training, helping avoid the underutilization of certain experts.
However, the classic \textit{k}\text{-}means algorithm lacks an inherent mechanism to balance cluster capacities, making balanced expert workloads difficult to achieve.
To address this challenge, TRIP incorporates capacity constraints into the \textit{k}\text{-}means clustering objective, ensuring that the size of each cluster $\mathcal{C}_m^k$ does not exceed a predefined target capacity $s_m$:
\begin{equation}
	\label{eq:cac}
	\quad \sum_{m=1}^{M} \sum_{\bm{x}_{ij}^k \in \mathcal{C}_m^k} \| \bm{x}_{ij}^k - \bm{\mu}_m^k \|^2 \quad \text{s.t.} \quad |\mathcal{C}_m^k| \leq s_m, \; \forall m,
\end{equation}
where
$s_m = \alpha \cdot \frac{N+1}{M}$, and $\alpha$ is a capacity factor hyperparameter.
To solve this constrained optimization problem, we introduce a Lagrange multiplier $\theta_m^k$ for each cluster and define the Lagrangian function:
\begin{equation}\label{eq:optimize_cac}
\mathcal{L} = \sum_{m=1}^{M} \left( \sum_{\bm{x}_{ij} \in \mathcal{C}_m^k} \| \bm{x}_{ij}^k - \bm{\mu}_m^k \|^2 + \theta_m^k (|\mathcal{C}_m^k| - s_m) \right),
\end{equation}
where
\( \theta_m^k \) acts as a flexible penalty term that adaptively regulates the assignment cost of tokens to each cluster.
When a cluster exceeds its capacity, its corresponding \( \theta_m^k \) increases, discouraging further token assignments.
Conversely, underutilized clusters have lower penalties, making them more likely to receive additional tokens. 

In addition to optimizing Eq.~\eqref{eq:optimize_cac} to softly enforce capacity constraints, we also implement a token replacement strategy to strictly maintain the condition $|\mathcal{C}_m^k| \leq s_m$. 
Specifically, when a cluster reaches its target capacity, a newly arriving token is accepted only if its assignment cost is lower than the current maximum cost within the cluster. If this condition is met, the token with the highest cost is discarded to maintain the capacity.
\subsection{Expert Assignment for Clusters}
Once the token clusters have been formed, the subsequent step involves assigning these clusters to experts.
Let the set of $M$ prompt experts at client $k$ be denoted as $\mathcal{E}^k = \{\bm{\mathcal{E}}_m^k\}_{m=1}^M$,
TRIP employs the OT algorithm to establish a optimal one-to-one correspondence between the $M$ clusters and the $M$ experts. 
Specifically, OT determines this correspondence by solving a constrained optimization problem that minimizes the overall transport cost between clusters and experts.
Formally, given a cost matrix \( \bm{\Lambda} \in \mathbb{R}^{M \times M} \), where each entry \( \Lambda_{i, j} \) represents the cost of assigning cluster $i$ to expert $j$,  OT seeks an optimal assignment  matrix \( \bm{\Gamma} \in \{0,1\}^{M \times M} \)  that minimizes the total allocation cost:
\begin{equation}\label{eq:OT}
	\min_{\bm{\Gamma}} \sum_{i=1}^{M} \sum_{j=1}^{M} \Lambda_{i, j} \Gamma_{i, j},
\end{equation}
subject to the marginal constraints:
\begin{equation}\label{eq:constraint}
	\sum_{j=1}^{M} \Gamma_{i, j} = 1, \sum_{i=1}^{M} \Gamma_{i, j} = 1,
\end{equation}
where $\Gamma_{i, j}=1$ indicates the assignment of cluster $i$ to expert $j$, and $\Gamma_{i, j}=0$ otherwise.
The constraints in Eq.~\eqref{eq:constraint} ensure that each cluster is assigned to exactly one expert and each expert is assigned exactly one cluster.

A intuitive approach for constructing the cost matrix $\bm{\Lambda}$ is to compute  semantic distance between cluster centroids and the prompt experts. However, as training progresses, prompt experts  continuously change, leading to instability in $\bm{\Lambda}$ and potentially inconsistent assignments of similar clusters across different images over time.
To tackle this challenge, TRIP introduces static, non-learnable keys, each associated with a prompt expert.
These keys remain fixed throughout training and serve as stable anchor points for cluster assignments.
The cost matrix \( \bm{\Lambda} \) is then computed based on the distance between cluster centroids and these static keys, rather than the dynamically evolving prompt experts:
\begin{equation}\label{eq:cost}
	\Lambda_{i, j} = 1 - \cos(\bm{\mu}_i^k, \bm{v}_j),
\end{equation}
where \( \bm{\mu}_i^k \) represents the centroid of the \( i \)-th token cluster at client \( k \),  \( \bm{v}_j \) is the static key  for the \( j \)-th prompt expert and $\cos(\cdot, \cdot)$ measures the cosine similarity. 
By computing the cost matrix based on static keys, semantically similar clusters---regardless of their originating image---are more likely to be assigned to the same expert.
This consistency encourages expert specialization, as each expert reliably receives clusters corresponding to specific semantic concepts.
Moreover, we randomly initialize these keys with orthogonality constraints, i.e., \( \bm{v}_i^\top \bm{v}_j = 0 \) for all \( i \neq j \).
This initialization strategy further promotes expert discriminability by minimizing inter-key correlations.
We employ the Hungarian algorithm~\cite{kuhn1955hungarian} to minimize the total cost encoded in the \( \bm{\Lambda} \), as it guarantees an optimal solution with a time complexity of $\mathcal{O}(M^3)$. Given that TRIP operates with a small number of clusters $M$, this computational cost is acceptable for our problem scale.

After the cluster-to-expert assignment for a given image $\bm{x}_i^k$, each prompt expert $\bm{\mathcal{E}}_i^k$ is associated with the tokens within a specific cluster.
The weight $\pi_i$ of $\bm{\mathcal{E}}_i^k$  is calculated as the ratio of tokens assigned to $\bm{\mathcal{E}}_i^k$ to the total number of tokens in all clusters. 
Consequently, the instance-specific prompt for the image $\bm{x}_i^k$ is generated by a weighted ensemble of the prompt experts:
\begin{equation}\label{eq:instance-p}
	\bm{E}_i^k = \quad \sum_{i=1}^{M} \pi_{i}  \bm{\mathcal{E}}_i^k.
\end{equation}
\subsection{Unbiased Prompt Expert Learning}
TRIP updates $M$ prompt experts  $\{\bm{\mathcal{E}}_m^k\}_{m=1}^M$ using domain-specific data on each client $k$.
Given a labeled sample from client $k$, i.e., $(\bm{x}^{k}_i,y^{k}_i) \in \mathcal{S}^k$, the instance-specific prompt $\bm{E}_i^k$ for image $\bm{x}_i^k$ can be derived using Eq.~\eqref{eq:instance-p}. 
The instance-specific prompt $\bm{E}_i^k$ and the description of category $c$ are then fed into the text encoder, generating the text feature $\bm{w}_{c,i}^k$ of category $c$, which is tailored to image $\bm{x}_i^k$. 
Similar to Eq.~\eqref{eq:clip}, 
the categorical probability distribution $\bm{p}^k(\bm{x}_i^k) \in [0,1]^C$, is computed as follows:
\begin{equation}\label{eq:p_k}
	p_c^k( \bm{x}_i^k ) = \frac{{\exp \left( {\cos ( {{\bm{w}_{c,i}^k},\bm{f}_i^k} )/\tau } \right)}}{{\sum\nolimits_{j = 1}^C {\exp \left( {\cos ({{\bm{w}_{j,i}^k},\bm{f}_i^k} )/\tau } \right)} }},
\end{equation}
where $\bm{f}_i^k$ is the image feature generated by the vision encoder for $\bm{x}_i^k$. Finally, the prompt experts can be optimized by minimizing the cross-entropy loss:
\begin{equation}\label{eq: s_ce}
	\mathcal{L}_{ce}^k =  - \frac{1}{|\mathcal{S}_k|}\sum\limits_{i = 1}^{|\mathcal{S}_k|} {\log \left( {p_{y_i^k}^k( {\bm{x}_i^k} )} \right)}.
\end{equation}

The cross-entropy loss $\mathcal{L}_{ce}^k$ is widely adopted in prompt learning-based FL methods to train domain-specific prompts for each client.  
However, optimizing this loss alone leads clients to train in complete isolation, causing the learned prompts to overfit to local data patterns.
As demonstrated in prior work~\cite{khattak2023self}, this specialization bias diminishes CLIP's fundamental capacity for cross-domain generalization. 
To address this issue, TRIP incorporates a debiasing strategy that leverages CLIP's zero-shot prediction distribution as a global reference. During the local training, each client minimizes the cross-entropy loss $\mathcal{L}_{ce}^k$  while simultaneously incorporating a Kullback-Leibler (KL) divergence loss to align its local prediction distribution with CLIP's zero-shot prediction distribution:
\begin{equation}\label{eq:kl_loss}
	{\cal L}_{kl}^k = \frac{1}{{|\mathcal{S}_k|}}\sum\limits_{i = 1}^{|\mathcal{S}_k|} {{p_c^g\left( {\bm{x}_i^k} \right)\log \left( {\frac{{p_c^g\left( {\bm{x}_i^k} \right)}}{{p_c^k\left( {\bm{x}_i^k} \right)}}} \right)} },
\end{equation}
where $p_c^g$ denotes CLIP's predicted probability  for category $c$ , and $p_c^k$ is the local predicted probability computed in Eq.~\eqref{eq:p_k}.
The optimization objective for client $k$ is thus defined as:
\begin{equation}\label{eq:total_loss}
	\mathcal{L}_{\text{total}}^k = \mathcal{L}_{ce}^k + \beta \mathcal{L}_{kl}^k,
\end{equation}
where $\beta$ is a hyperparameter that balances the two losses.

Once local training is finished, each client uploads its prompt experts to the server. The server then aggregates these experts globally using the FedAvg~\cite{mcmahan2017communication} algorithm, with aggregation weights proportional to the size of each client's dataset:
\begin{equation} \label{eq:agg}
\bm{\mathcal{E}}_i^g  = \sum\limits_{k = 1}^{K} \frac{|\mathcal{S}_k|} {|\mathcal{S}|}\bm{\mathcal{E}}_i^k.
\end{equation}
The aggregated prompt experts are subsequently distributed back to the clients for the next training iteration. For clarity, the complete training process of TRIP is presented in Algorithm~\ref{alg:training}.

\begin{algorithm}[t]
	\caption{The training process of TRIP.}
	\label{alg:training}
	\begin{algorithmic}[1]
		\Statex \textbf{Input:} Datasets of $K$ clients $\{\mathcal{S}^k\}_{k=1}^{K}$, 
		CLIP model, total communication rounds $R$, 
		and total local epochs $E$.
		\Statex \textbf{Output:} Global prompt expert set $\mathcal{E}^g =  \{\bm{\mathcal{E}}_m^g\}_{m=1}^M$.
		\State \textbf{For} $r = 1$ to $R$ \textbf{do}
		\vspace{-0.2em}
		\Statex \begin{ServerBlock}
			\State \hspace*{1em} \textbf{Server:} Send keys $ \{\bm{v}_m\}_{m=1}^M$ to all clients only once.
			\State \hspace*{1em} Send the global prompt expert set $\mathcal{E}^g$ to all clients.
		\end{ServerBlock}
		\Statex \begin{ClientBlock}
			\State \hspace*{1em} \textbf{Client:} Initialize $\mathcal{E}^k \gets \mathcal{E}^g$.
			\State \hspace*{1em} \textbf{For} $e = 1$ to $E$
			\State \hspace*{2em} \textbf{For each} $\bm{x}_i^k \in \mathcal{S}^k$
			\State \hspace*{3em} Extract token embeddings of $\bm{x}_i^k$ using CLIP.
			\State \hspace*{3em} Generate token clusters by optimizing Eq.~\eqref{eq:optimize_cac}.
			\State \hspace*{3em} Assign clusters to experts via OT and $\{\bm{v}_m\}_{m=1}^M$.  
			\State \hspace*{3em} Generate the prompt $\bm{E}_i^k$ for $\bm{x}_i^k$ using Eq.~\eqref{eq:instance-p}.
			\State \hspace*{3em} Update the prompt experts in ${\mathcal{E}}^k$ using Eq~\eqref{eq:total_loss}.
			\State \hspace*{2em} \textbf{End For}
			\State \hspace*{1em} \textbf{End For}
			\State \hspace*{1em} Upload $\mathcal{E}^k$ to the server.
		\end{ClientBlock}
		\Statex \begin{ServerBlock}
			\State \hspace*{1em} \textbf{Server:} Update the global experts in $\mathcal{E}^g$ using Eq.~\eqref{eq:agg}.
		\end{ServerBlock}
		\State \textbf{End For}
	\end{algorithmic}
\end{algorithm}

\section{Experiments}\label{sec:experiments}
In this section, we provide a detailed description of the experimental settings and present representative results to validate the effectiveness of our proposed method.
\subsection{Datasets and Evaluation Protocol}
Our experiments are conducted on four widely used datasets. Brief introductions to these datasets are as follows.
\begin{itemize}
	
	\item \textbf{PACS}~\cite{li2017deeper} spans four distinct domains: Photo, Sketch, Cartoon, and Art Paintings,  comprising 9,991 images distributed across 7 classes.

	\item \textbf{Office-Home}~\cite{venkateswara2017deep}  consists of 24,788 images categorized into 65 classes. This dataset includes four domains: Art, Clipart, Product, and Real World. 
	
	\item \textbf{VLCS}~\cite{fang2013unbiased} comprises 10,729 instances across 5 classes, distributed among four domains: VOC2007, LabelMe, Caltech-101, and SUN09.
	
	\item \textbf{DomainNet}~\cite{peng2019moment} is a large-scale dataset containing 569,010 images across 345 classes, spanning six photographic domains: Clipart, Infograph, Painting, Quickdraw, Real World, and Sketch.
\end{itemize}

We adopt the standard \texttt{leave-one-domain-out} evaluation protocol~\cite{zhang2023federated} for all datasets. In this setup, one domain is chosen as the unseen target domain, while the remaining domains are distributed across clients as source domains for federated training. 
The data within each source domain are split into training and evaluation sets, keeping the same split ratio as~\cite{gong2024federated,huang2020self,xu2021fourier} for PACS, Office-Home, and VLCS, and~\cite{peng2019moment} for DomainNet. 
We report classification accuracy for each target domain, along with the average accuracy across all domains for every dataset.
 
\subsection{Baselines}
We compare TRIP with 15 recently proposed methods, across four categories:

\emph{1) Centralized learning (CL)-based DG methods}
leverage aggregated data from all source domains in a centralized manner during training. 
\textbf{SWAD}~\cite{cha2021swad} is a classic DG algorithm designed to identify flat minima. It provides a theoretical analysis demonstrating that convergence to flatter minima effectively reduces the domain generalization gap.
\textbf{HCVP}~\cite{zhou2024hcvp} proposes a hierarchical contrastive visual prompt learning framework, which jointly optimizes domain-specific and task-specific visual features to achieve superior generalization performance.
\textbf{Doprompt}~\cite{zheng2022prompt}  utilizes visual prompt tuning to enhance generalization. It dynamically assigns weights to domain-specific prompts for each target image.

\emph{2) FL-based methods} 
enable multiple decentralized clients to collaboratively train a shared global model while preserving data privacy by eliminating the need for direct data sharing.
\textbf{FedAvg}~\cite{mcmahan2017communication} is the most fundamental FL algorithms, which averages the parameters of local models to generate the global model.
\textbf{FedProx}~\cite{li2020federated} is designed to address statistical heterogeneity among client data. It incorporates a proximal regularization term that constrains the local model to remain close to the global model.

\emph{3) Conventional FedDG methods} 
aim to build a well-generalized model for unseen target domains within a FL framework.
\textbf{FedSR}~\cite{nguyen2022fedsr} employs the $L_2$-norm and conditional mutual information regularizers to learn simple yet essential data representations.
\textbf{FedADG}~\cite{zhang2021federated} seeks to extract universal feature representations by adversarially aligning source domain data distributions with an adaptively generated reference distribution.
\textbf{CCST}~\cite{chen2023federated} facilitates the sharing of domain-specific styles across clients, which allows local models to generalize across diverse image styles from all domains, thereby mitigating model bias induced by local data distributions.
\textbf{ELCFS}~\cite{liu2021feddg} enhances model generalizability by leveraging information exchange among clients in the frequency domains.
\textbf{GA}~\cite{zhang2023federated} incorporates a variance reduction regularizer based on generalization gaps, which dynamically adjusts the
aggregation weights of local models.
\textbf{StableFDG}~\cite{park2024stablefdg} enhances generalization performance through a two-stage framework. It first performs cross-client style sharing, and then incorporates an attention-based highlighter to extract domain-invariant features across different styles.

\emph{4) Parameter-efficient fine-tuning (PEFT)-based methods} 
freeze the parameters of  VLMs and optimize only a small number of additional parameters, such as learnable prompt tokens or adapters.
\textbf{FedCLIP}~\cite{lu2023fedclip} incorporates a lightweight adapter to customize VLMs to individual clients.
\textbf{PromptFL}~\cite{guo2023promptfl} learns text prompts locally and aggregates them into a global prompt via the FedAvg algorithm.
\textbf{FedOTP}~\cite{li2024global} adopts the OT algorithm to align  consensus-aware global prompts with  category-aware local prompts, thereby achieving a balance between global generalization and local personalization.
\textbf{FedAPT}~\cite{su2024federated} incorporates domain-specific information into prompts through an adaptive module, generating personalized text prompts for test samples.
\subsection{Implementation Details}
Our method was implemented based on the CLIP model, with the ViT-Base/16 backbone selected as the image encoder. 
For federated training, we set the number of local update epochs to 1 and the total number of communication rounds to 15. We adopted the AdamW optimizer with a learning rate of $4 \times 10^{-4}$ and a batch size of 16. The capacity factor $\alpha$ in Eq.~\eqref{eq:optimize_cac} was set to 1.0 during training and was increased to 2.0  for inference.   The balancing  coefficient $\beta$ in Eq.~\eqref{eq:total_loss} was set to 0.8. 
All experimental configurations remained consistent across different datasets.
To facilitate reproducibility, our code is publicly available at~\url{https://github.com/GongShuai8210/TRIP}.

For the reproduced baseline methods, we cited the results of CL-based DG methods with a ViT~\cite{dosovitskiy2020image} backbone as reported in prior study~\cite{zhou2024hcvp}. For all other baseline categories, we reimplemented the methods using CLIP to ensure a fair comparison. Hyperparameters for these methods were configured according to either the original papers or performance on the validation set. 
Additionally, 
the token-level and image-level learnable routers presented in Subsection~\ref{subsec:image-level router} are implemented using a single-layer FNN followed by a softmax function for expert assignment, and are optimized according to the objectives proposed in prior work~\cite{fedus2022switch}.
\subsection{Overall Performance}\label{subsec:performance}

We report results for two configurations of TRIP. 
The first is the standard TRIP, which employs four prompt experts with 32 prompt tokens per expert. 
The second configuration, referred to as TRIP-Lite, represents a minimal-parameter setting designed to minimize optimization effort, using only two prompt experts with one token each.
All experimental results represent the average of three runs, with the best performance for each generalization task emphasized in bold. 
Additionally, we provide the communication cost per round in the final columns of Tables~\ref{pacs_comparasion}--\ref{domainnet_comparasion} for comparison.
\begin{table}[t]
	\begin{center}
		\caption{Performance comparison in accuracy (\%) on PACS.}\label{pacs_comparasion}
		\resizebox{0.48\textwidth}{!}{
			\begin{tabular}{c|c|c|c|c|c|c}
					\toprule
					\multirow{2}{*}{Methods}  & \multicolumn{6}{c}{PACS}\\
					\cmidrule{2-7} 
					& {Art} & {Cartoon} & {Photo} & {Sketch} & {Avg.}  &{Com.cost}\\
					\midrule
					\multicolumn{7}{>{\columncolor{LightBlue}}c}{\textit{CL-based DG methods}}\\
					
					SWAD &93.23 &85.93 &99.18 &82.03&90.44&0M \\
					HCVP &93.17 &86.89 &99.33 &81.30  &90.17&0M \\ 
					Doprompt &95.00  &86.35  &99.63  &78.20  &89.91&0M \\
					\midrule
					\multicolumn{7}{>{\columncolor{LightOrange}}c}{\textit{FL-based DG methods}}\\
					FedAvg   & 82.98 &	65.64 &	97.72 &	65.29 &	77.91 &149.62M \\
					FedProx  & 84.33 &	68.52 &	97.80 &	64.66 &	78.83 &149.62M \\
					\midrule
					\multicolumn{7}{>{\columncolor{LightGreen}}c}{\textit{Conventional FedDG methods}}\\
					FedSR   &88.39 	&67.66 	&95.90 	&66.17 	&79.53 &149.62M \\
					FedADG   &83.25 	&65.72 	&98.30 	&65.68 	&78.24 &150.14M\\
					CCST   	 &87.15 	&74.96 	&98.58 	&66.06 	&81.69 &149.62M \\
					ELCFS   &87.05 	&73.58 	&98.32 	&65.31 	&81.07 &149.62M\\
					ELCFS+GA  &87.94 	&75.52 	&97.79 	&65.98 	&81.81 &149.62M \\
					StableFDG 	&91.12 	&78.57 	&98.11 	&66.83 	&83.66 	 &151.55M \\ 
					\midrule
					\multicolumn{7}{>{\columncolor{LightRed}}c}{\textit{PEFT-based methods}}\\
					FedCLIP   	&96.21 	&98.92 	&99.70 &85.97 &95.20 &0.524M\\
					PromptFL &96.41 &98.43 	&99.62 	&91.21 	&96.41 & 0.008M\\
					FedOTP  &96.46	&98.67	&99.77&	\textbf{91.34}&96.56&0.008M\\
					FedAPT  &97.26 	&98.59	&99.71 	&90.52	&96.52 &0.059M\\
					\textbf{TRIP-Lite (Ours)}  &97.43	&98.62	&99.80 	&90.45	&96.58 &0.001M\\
					\textbf{TRIP (Ours)} &\textbf{98.14}	&\textbf{99.19}	&\textbf{99.88}	&90.37&	\textbf{96.90}&0.065M\\
					\bottomrule
				\end{tabular}
			}
		\end{center}
		
	\end{table}		
	\begin{table}[t]
		\vspace{-0.8pt}
		\begin{center}
			\caption{Performance comparison in accuracy (\%) on Office-Home.}\label{officehome_comparasion}
			
			\resizebox{0.48\textwidth}{!}{
				\begin{tabular}{c|c|c|c|c|c|c}
						\toprule
						\multirow{2}{*}{Methods}  & \multicolumn{6}{c}{Office-Home}\\
						\cmidrule{2-7} 
						& {Art} & {Clipart} & {Product} & {Real} & {Avg.}  &{Com.cost}\\
						\midrule
						\multicolumn{7}{>{\columncolor{LightBlue}}c}{\textit{CL-based DG methods}}\\
						
						SWAD &76.26 &68.87 &86.74 &87.03 &79.73&0M \\
						HCVP &81.77 &69.76 &88.01 &90.62&82.54 &0M\\ 
						Doprompt  &80.95 &70.88 &88.94 &90.10 &82.72&0M \\
						\midrule
						\multicolumn{7}{>{\columncolor{LightOrange}}c}{\textit{FL-based DG methods}}\\
						FedAvg   &69.68 	&48.42 &72.70 	&78.85 	&67.41 &149.62M \\
						FedProx   	&71.16 	&49.10 	&72.51 	&78.71 	&67.87 &149.62M \\
						\midrule
						\multicolumn{7}{>{\columncolor{LightGreen}}c}{\textit{Conventional FedDG methods}}\\
						FedSR   &69.59 	&50.01 	&72.98 	&79.45 	&68.01 &149.62M \\
						FedADG   &69.75 	&49.21 	&73.03 	&79.31 	&67.83 &150.14M\\
						CCST   	 &69.54 	&51.59 	&72.24 	&81.49 	&68.72 &149.62M \\
						ELCFS   &69.39 	&50.68 	&71.91 	&80.59 	&68.14 &149.62M\\
						ELCFS+GA  &68.68 	&52.32 	&73.71 	&81.60 	&69.08 &149.62M \\
						StableFDG  &69.51 	&53.21 	&73.29 	&82.01 	&69.51  &151.55M \\
						\midrule
						\multicolumn{7}{>{\columncolor{LightRed}}c}{\textit{PEFT-based methods}}\\
						FedCLIP   	&78.45	&64.77 &87.68&	87.84&	79.69&0.524M\\
						PromptFL   &80.75	&69.88 &90.96&	90.25&	82.96& 0.008M\\
						FedOTP  &81.16	&70.23	&90.98&	90.82&	83.30&0.008M\\
						FedAPT  &82.96	&70.98 &91.33 &90.51 &83.94&0.534M\\
						\textbf{TRIP-Lite (Ours)} &84.21	&70.19	&92.34	&91.41&	84.53&0.001M\\
						
						\textbf{TRIP (Ours)} &\textbf{85.49}	&\textbf{71.22}	&\textbf{92.97}		&\textbf{91.79}&	\textbf{85.37}&0.065M\\
						\bottomrule
					\end{tabular}
				}
			\end{center}
		\end{table}		
		
Table~\ref{pacs_comparasion} shows the comparison results on PACS. 
We can see that CL-based DG methods exhibit superior performance compared to FL-based DG methods and conventional FedDG methods. 
Even DoPrompt, the lowest-performing CL-based model, surpasses the state-of-the-art (SOTA) conventional FedDG method, StableFDG, by a large margin of 6.25\% in average accuracy.
However, the performance improvement is obtained by compromising data privacy. 
The results of FL-based DG methods significantly underperform conventional FedDG methods, 
likely because they focus on adapting to local client data rather than generalizing to unseen domains. 
Among conventional FedDG methods, CCST, ELCFS, and StableFDG, which exchange image style information to facilitate knowledge sharing among clients, have demonstrated better performance than FedSR and FedADG.  
PEFT-based methods achieve superior performance and communication efficiency, markedly surpassing the other three categories of methods.
Our method, TRIP,  achieves the highest average accuracy, outperforming the second-best method, FedAPT, by 0.38\%. Its lightweight variant, TRIP-Lite, achieves comparable performance to FedAPT while reducing communication parameters 59 times, from 0.059M to 0.001M.

Table~\ref{officehome_comparasion} presents the comparison results on Office-Home. 
TRIP demonstrates a clear advantage on this dataset, achieving the best results across all evaluated domains.
Compared to the previous SOTA method, FedAPT, TRIP achieves an average performance improvement of 1.43\% on this dataset. Meanwhile, it reduces the number of communication parameters from 0.534M to 0.065M, representing a communication cost reduction to just 12.2\%.
The TRIP-Lite version further reduces the communication parameters by 99.8\%  while still maintaining an average performance improvement of 0.59\%.

Table~\ref{vlcs_comparasion} displays the comparison results on VLCS. 
We observe that  TRIP demonstrates the optimal performance in three out of the four domains, with an average improvement  of 0.99\% over prior SOTA results, FedAPT. 
Especially in the VOC domain, TRIP surpasses FedAPT by 1.01\% and is the only method to exceed 81\% accuracy. 
TRIP-Lite still demonstrates its communication efficiency compared to FedAPT, reducing the communication parameters by a factor of 43, while achieving a 0.19\% improvement in average accuracy. 

Table~\ref{domainnet_comparasion} shows the comparison results on DomainNet.
Due to the prohibitive computational costs of reproducing all models on this large-scale dataset, we restrict our evaluation to PEFT-based methods and the 16-shot setting. 
Our method, TRIP, obtains the best results across all six domains, achieving an average improvement of 2.02\%, with communication  cost reduction to 43 fewer than FedAPT.
Furthermore, requiring only 0.001M communication parameters—just $3.5 \times 10^{-4}$ the size of FedAPT—TRIP-Lite achieves a 0.48\% performance improvement, highlighting its effectiveness.
\begin{table}[t]
	\begin{center}
		\caption{Performance comparison in accuracy (\%) on VLCS.}\label{vlcs_comparasion}
		\resizebox{0.48\textwidth}{!}{
			\begin{tabular}{c|c|c|c|c|c|c}
					\toprule
					\multirow{2}{*}{Methods}  & \multicolumn{6}{c}{VLCS}\\
					\cmidrule{2-7} 
					& {Caltech} & {LabelMe} & {VOC} & {SUN} & {Avg.}  &{Com.cost}\\
					\midrule
					\multicolumn{7}{>{\columncolor{LightBlue}}c}{\textit{CL-based DG methods}}\\
					SWAD & 98.49 &68.36 &75.40  &79.49  &79.31&0M \\
					HCVP &96.32  &66.26  &80.08  &81.65  &81.08&0M  \\
					Doprompt &96.70  &66.53  &78.28  &79.39  &80.23&0M \\
					\midrule
					\multicolumn{7}{>{\columncolor{LightOrange}}c}{\textit{FL-based DG methods}}\\
					FedAvg   &95.80 	&66.07 	&78.64 	&73.56 	&78.52 &149.62M \\
					FedProx   &95.56 	&65.10 	&76.85 	&78.26 	&78.94 &149.62M \\
					\midrule
					\multicolumn{7}{>{\columncolor{LightGreen}}c}{\textit{Conventional FedDG methods}}\\
					FedSR   &95.86 	&66.01 	&78.79 	&73.85 	&78.63 &149.62M \\
					FedADG   &95.41 	&65.92 	&77.64 	&76.33 	&78.83 &150.14M\\
					CCST   	 &96.98 	&66.01 	&76.80 	&77.78 	&79.39 &149.62M \\
					ELCFS   &95.97 	&65.32 	&76.84 	&78.19 	&79.08& 149.62M\\
					ELCFS+GA  &97.13 	&65.60 	&79.13 &79.28 	&80.29 &149.62M \\
					StableFDG 	&97.79 	&66.13 	&80.12 	&79.56 	&80.90&151.55M \\
					
					\midrule
					\multicolumn{7}{>{\columncolor{LightRed}}c}{\textit{PEFT-based methods}}\\
					FedCLIP   &99.93	&66.98	&73.28	&87.14	&81.83&0.524M\\
					PromptFL   &99.65	&68.03	&72.24	&85.10&	81.26& 0.008M\\
					FedOTP  &99.46	&\textbf{68.42}	&72.65&	85.26&	81.45&0.008M\\
					FedAPT  &99.29	&67.05	&80.70&	85.86&83.22&0.043M\\
					\textbf{TRIP-Lite (Ours)} &99.92	&66.82	&80.77	&86.13&	83.41&0.001M\\
					\textbf{TRIP (Ours)} &\textbf{99.93}	&67.73	&\textbf{81.71}	&\textbf{87.47}&	\textbf{84.21}&0.065M\\
					\bottomrule
				\end{tabular}
			}
		\end{center}
	\end{table}	
\subsection{Ablation Study}
\begin{table}[t]
	\begin{center}
		\caption{Performance comparison in accuracy (\%) on DomainNet.}\label{domainnet_comparasion}
		\vspace{-10pt}
		\resizebox{0.49\textwidth}{!}{
			\begin{tabular}{c|c|c|c|c|c|c|c|c}
					\toprule
					\multirow{2}{*}{Methods}  & \multicolumn{8}{c}{DomainNet}\\
					\cmidrule{2-9} 
					& {Clipart} & {Info} & {Paint} & {Quick} & {Real}&{Sketch}&{Avg.}&{Com.cost}\\
					\midrule
					\multicolumn{9}{>{\columncolor{LightRed}}c}{\textit{PEFT-based Methods}}\\
					FedCLIP  &71.58 	&46.25 	&65.42 &	14.91 &	82.57 &	61.03 &	56.96 
					&0.524M\\
					PromptFL  &72.07 &	44.62 &	64.13 &	11.74 &	81.29 &	63.12 &	56.16 
					&0.008M\\
					FedOTP  &72.59 &	45.68 &	65.76 &	12.97 &	82.69 &	64.26 &	57.33 
					&0.008M\\
					FedAPT &72.82 	&47.98 	&67.09 	&13.25 	&82.70 	&64.98 &	58.14 
					&2.829M \\
					\textbf{TRIP-Lite (Ours)} &73.05 &	48.90 &	67.65 &	14.13 &	82.32 &	65.67 &	58.62 &0.001M\\
					\textbf{TRIP (Ours)} &\textbf{73.70}	&\textbf{51.92}	&\textbf{69.65}	&\textbf{15.58} 	&\textbf{84.06 }	&\textbf{66.04 }	&\textbf{60.16} &0.065M\\
					\bottomrule
				\end{tabular}
			}
		\end{center}
	\end{table}
\begin{table}[t]
	\vspace{-1.2pt}
	\centering
	\caption{Investigation on the contribution of different components.}
		\vspace{1.3pt}
	\label{tab:ablation}
	\resizebox{0.42\textwidth}{!}{
		\begin{tabular}{c|c|c|c|c|c|c}
			\toprule
			Capacity & Keys & Cluster &$\mathcal{L}_{kl}^k$& PACS  & Office-Home & VLCS \\ 
			\midrule
			$\checkmark$ & $\checkmark$ & $\checkmark$ & $\checkmark$ & \textbf{96.90}& \textbf{85.37} & \textbf{84.21}  \\
			$\boldsymbol{\times}$ & $\checkmark$ & $\checkmark$ & $\checkmark$ & 96.67 & 85.02 & 83.51\\
			$\boldsymbol{\times}$ & $\boldsymbol{\times}$ & $\checkmark$ & $\checkmark$ & 96.55 & 84.65  & 82.76  \\

			$\boldsymbol{\times}$ & $\boldsymbol{\times}$ & $\boldsymbol{\times}$ & $\checkmark$  & 96.62 & 84.54  & 82.32  \\
			$\boldsymbol{\times}$ & $\boldsymbol{\times}$ & $\boldsymbol{\times}$ & $\boldsymbol{\times}$  & 96.38 & 84.32  & 82.01   \\			
			\bottomrule
		\end{tabular}
	}
\end{table}
Table~\ref{tab:ablation} presents the accuracy of TRIP when key components are selectively omitted. As described in Eq.~\eqref{eq:cac}, we introduce the capacity constraint to balance token distribution among clusters.
Because TRIP assigns all tokens within a cluster to a single expert, this constraint inherently enforces balanced expert workloads. 
When the capacity constraint is removed, the model exhibits a slight performance decline, particularly on  VLCS, where accuracy drops by 0.70\%.
This performance degradation likely arises from imbalanced expert workloads, which leads to insufficient training of certain experts.

TRIP employs static keys to ensure that semantically similar tokens are assigned to the same expert. To evaluate the effectiveness of this design, we  replace the static keys with dynamically updated prompt experts. 
The results show a
further performance decline, with accuracy dropping by
0.12\%, 0.37\%, and 0.75\% on the PACS, Office-Home, and
VLCS, respectively.
This decline occurs because dynamic prompt experts fail to preserve the semantic consistency of token-to-expert assignments, particularly for tokens from different images, thereby compromising expert specialization.

After ablating both aforementioned components, we further eliminate the token clustering process and implement a random token-to-expert assignment strategy.  
The empirical results demonstrate that random assignment induces performance instability. The model exhibits performance degradation on both Office-Home and VLCS, while unexpectedly achieving performance gains on PACS.
This behavior is likely due to the random assignment strategy preventing experts from effectively specializing.

In TRIP, the prompt experts are trained locally, using KL divergence loss $\mathcal{L}_{kl}^k$  to align their predicted distribution with CLIP's zero-shot inference distribution.
Our experiments show that removing $\mathcal{L}_{kl}^k$ results in substantial performance degradation, with accuracy decreases of 0.24\%, 0.22\%, and 0.31\% on  PACS, Office-Home, and VLCS, respectively. 
This drop in performance might arise from the prompt experts overfitting to local data patterns, thereby reducing the generalization ability.

\subsection{Comparison of Image-Level and Token-Level Router}\label{subsec:image-level router}
\begin{figure}[t]
	\centering
	\includegraphics[width=1\linewidth]{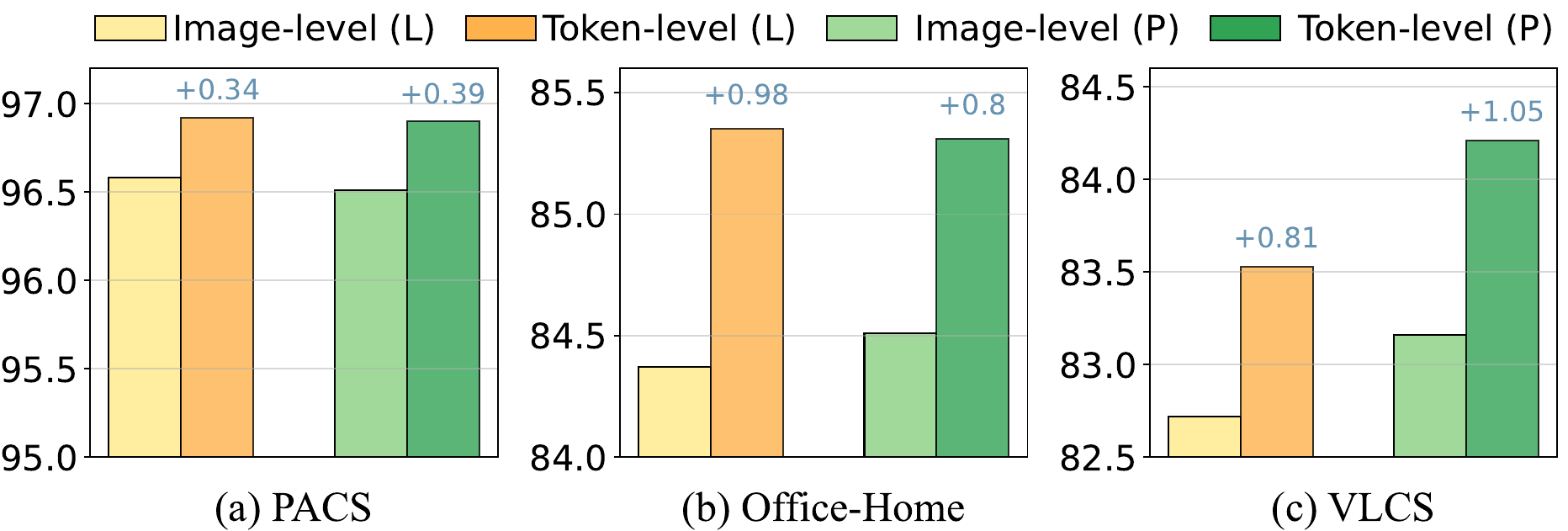}
	\captionsetup{skip=-1pt}
	\caption{Comparison of learnable (L) and our parameter-free (P) image-level and token-level router designs across different datasets}
	\label{fig:router_design}
\end{figure}
In this section, we investigate the impact of token-level and image-level routers on model performance. 
We evaluate TRIP and its three variants: 1) a parameter-free image-level router which clusters images within a mini-batch and assigns images from the same cluster to the same prompt expert; 2) a learnable image-level router which assigns images to experts based on global image features; and 3) a learnable token-level router which assigns image tokens to experts based on token embeddings.
As illustrated in Fig.~\ref{fig:router_design}, the token-level routers consistently outperform their image-level counterparts, regardless of whether their designs are learnable or parameter-free. 
This advantage may stem from the token-level approach’s ability to enable distinct experts to capture fine-grained image information.
Notably, our parameter-free method  achieves competitive performance compared to learnable routing approaches, demonstrating its advantages in both parameter efficiency and generalization capability.
\subsection{Effect of Key Initialization}
\begin{table}[t]
	\centering
	\caption{Effect of various key initialization strategies}
	\label{tab:init_keys}
	\resizebox{0.38\textwidth}{!}{
		\begin{tabular}{c|c|c|c|c}
			\toprule
			Initialization & PACS & Office-Home & VLCS &Avg. \\ 
			\midrule
			Uniform & 96.50 & 85.00 & 83.67 & 88.39\\
			Normal& 96.80 &85.14  & 83.69 & 88.54\\
			Binary & 96.83 & 85.07 & 83.50 & 88.47 \\
			Orthogonal & \textbf{96.90} & \textbf{85.37} & \textbf{84.21} & \textbf{88.83} \\
			\bottomrule
		\end{tabular}
	}
\end{table}
TRIP utilizes static keys to facilitate stable token-to-expert assignment, where each key is initialized to be mutually orthogonal to the others.
To evaluate the effect of different initialization methods on model performance, we conduct experiments with four strategies: (1) uniform initialization, where elements are randomly sampled from the interval $[0, 1]$; (2) normal initialization using the standard normal distribution $\mathcal{N}(0, 1)$; (3) binary initialization, where elements are randomly set to 0 or 1; and (4) orthogonal initialization, which ensures mutual orthogonality among keys.  As presented in Table~\ref{tab:init_keys}, among all methods, orthogonal initialization achieves the best performance, while the other methods yield comparable results. 
This effectiveness can be attributed to orthogonal initialization enhancing expert discriminability, thereby facilitating more accurate token-to-expert assignment.
\subsection{Analysis of Model Inference Stability}
\begin{figure}[t]
	\captionsetup[subfloat]{labelfont={scriptsize},textfont={scriptsize}}
	\centering
	\subfloat[Office-Home]{
		\includegraphics[width=0.8\linewidth]{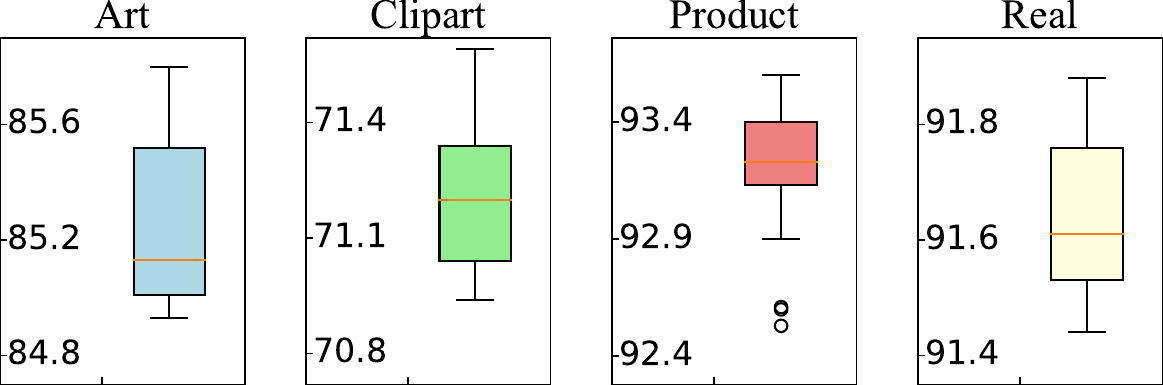}
		\label{infer_office}
	}\\
	\subfloat[VLCS]{
		\includegraphics[width=0.8\linewidth]{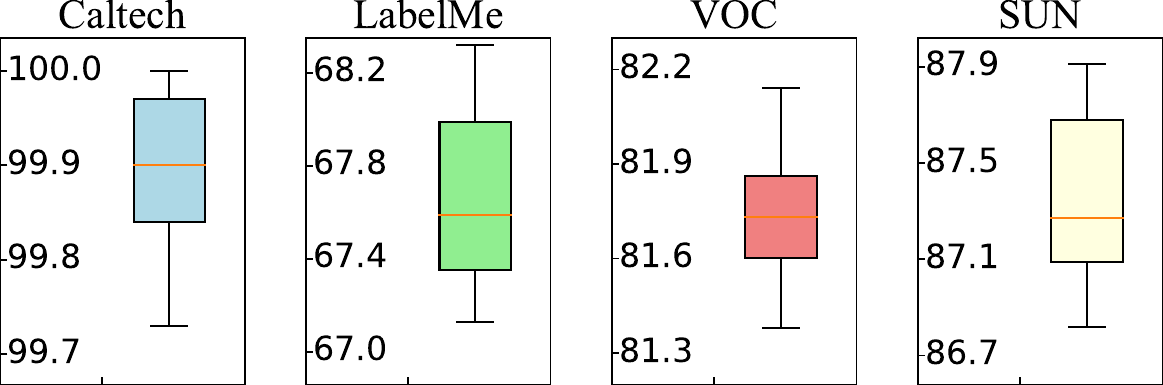}
		\label{infer_vlcs}
	}
	\caption{Visualization of model inference stability under different random seeds on Office-Home and VLCS.}
	\label{fig:infer}
\end{figure}
TRIP employs a clustering-based routing mechanism to avoid introducing additional trainable parameters. Unlike learnable routers, which produce deterministic routing decisions once trained, clustering-based methods can yield varying results if the clustering process is re-initialized during inference, which might lead to inference instability.
We therefore visualized the variation in model performance across different random seeds using box plots.
As shown in Fig.~\ref{fig:infer}, the model performance exhibits minor fluctuations within a narrow range, demonstrating robust stability across multiple domains. 
The observed stability arises from static-key anchoring of the prompt experts, allowing  different clusters to be reliably assigned to their corresponding experts.
\subsection{Hyperparameter Analysis}
\subsubsection{Effect of Capacity Factor}
\begin{figure}[t]
	\centering
	\includegraphics[width=0.9\linewidth]{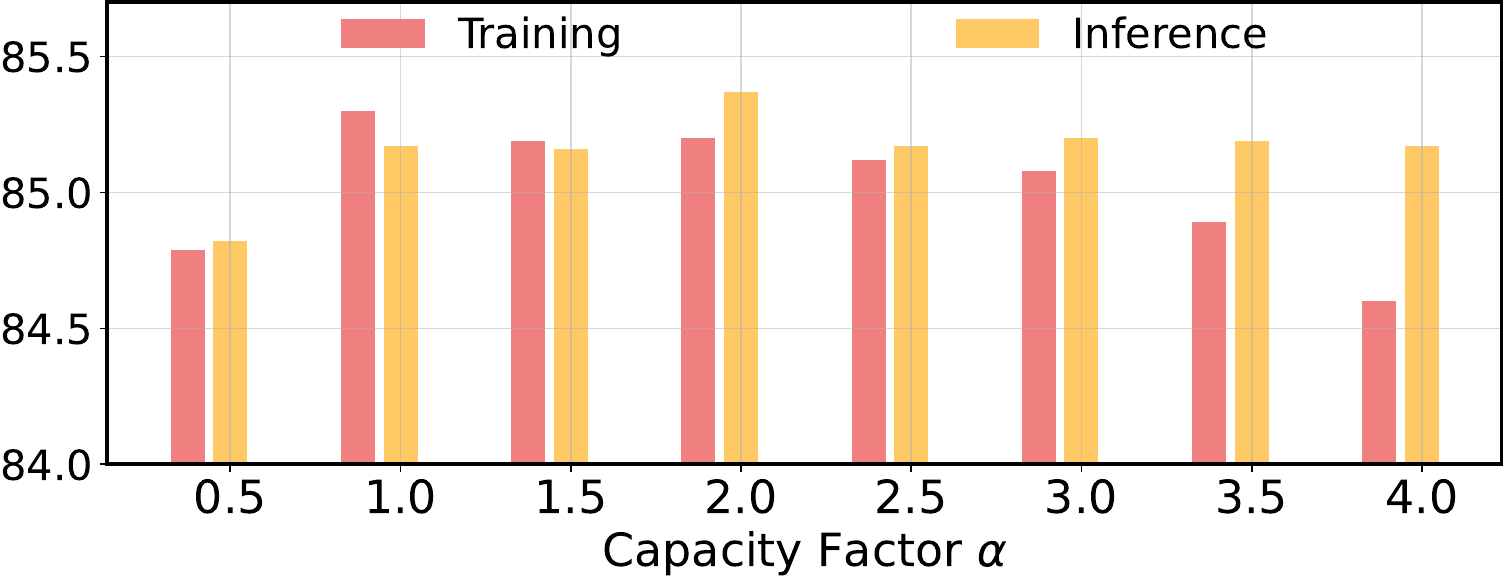}
	\captionsetup{skip=-1pt}
	\caption{Effect of capacity factor $\alpha$ during training and inference.}
	\label{fig:ca}
\end{figure}
Fig.~\ref{fig:ca} presents the effect of the capacity factor $\alpha$ on model performance. During training, the model performance increases from $\alpha = 0.5$ to $1.0$, but declines with further increases in $\alpha$. This trend occurs because a small $\alpha$ leads to more tokens being dropped during clustering, resulting in incomplete image information being processed, whereas an excessively large $\alpha$ induces expert workload imbalance, hindering effective expert learning.

After expert learning is complete, we follow the practice proposed in Switch Transformer~\cite{fedus2022switch} and re-evaluate the effect of $\alpha$ on inference performance. 
The results show that using a capacity factor larger than the training-optimal value (i.e., $\alpha = 1.0$) leads to improved performance, peaking at $\alpha = 2.0$. 
This finding is consistent with observations reported in previous work~\cite{fedus2022switch}. 
This improvement is attributed to a larger capacity factor reducing the number of dropped tokens. At the same time, the potential negative impact of imbalanced expert workloads induced by this factor becomes negligible, as the experts have already been sufficiently trained.
\subsubsection{Effect of Loss Weight}
\begin{table}[t]
	\centering
	\captionsetup{position=above}
	\caption{Effect of the KL loss coefficient $\beta$.}
	\resizebox{0.38\textwidth}{!}{
		\begin{tabular}{l|c|c|c|c|c}
			\toprule
			$\beta$          & Art & Clipart & Product & Real & Avg. \\
			\midrule
			0.02         & 83.93   & 71.75 &92.77 &90.91 &84.84\\
			0.2          & 84.46   & \textbf{71.91} &92.72 &91.46 &85.14\\
			0.4          & 84.87   & 71.01 &92.85 &91.75 &85.12\\
			0.6          &85.00    &71.13  &92.88 &91.78 &85.20 \\
			0.8			&\textbf{85.49}  &71.22 &\textbf{92.97} &\textbf{91.79} &\textbf{85.37}     \\              
			1    		 &85.24	   &71.24  &92.65 &91.50 &85.16        \\
			10    		 &84.75	   &68.86  &90.80 &90.24 &83.66        \\
			\bottomrule
		\end{tabular}
	}
	\label{tab:weights}
\end{table}
The KL loss coefficient $\beta$ in Eq.~\eqref{eq:total_loss} balances the task-specific objective and regularization in optimizing the prompt experts.
Table~\ref{tab:weights} reports the results of TRIP across different values of $\beta$.
The results improve progressively as $\beta$ increases from 0.02, reaching a peak at 0.8,
highlighting the importance of KL loss in TRIP.
However, with further increases in $\beta$, performance gradually declines. This occurs because an excessively large $\beta$ causes the KL loss to dominate the training process, thereby hindering the experts' ability to capture local data patterns.
\subsubsection{Effect of  Number of Prompt Tokens and Experts}
\begin{figure}[t]
	\captionsetup[subfloat]{labelfont={scriptsize},textfont={scriptsize}}
	\centering
	\subfloat[The number of prompt tokens ]{\includegraphics[width=0.5\linewidth]{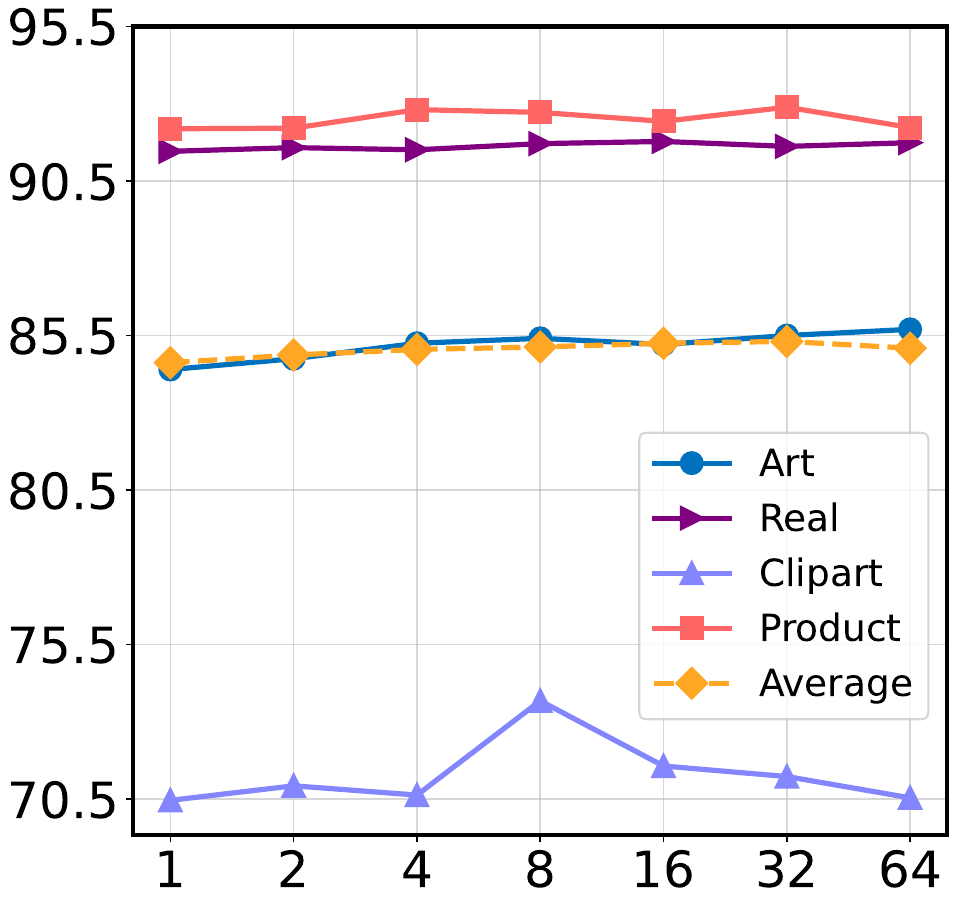}
		\label{p_length}}
	\subfloat[The number of prompt  experts]{\includegraphics[width=0.5\linewidth]{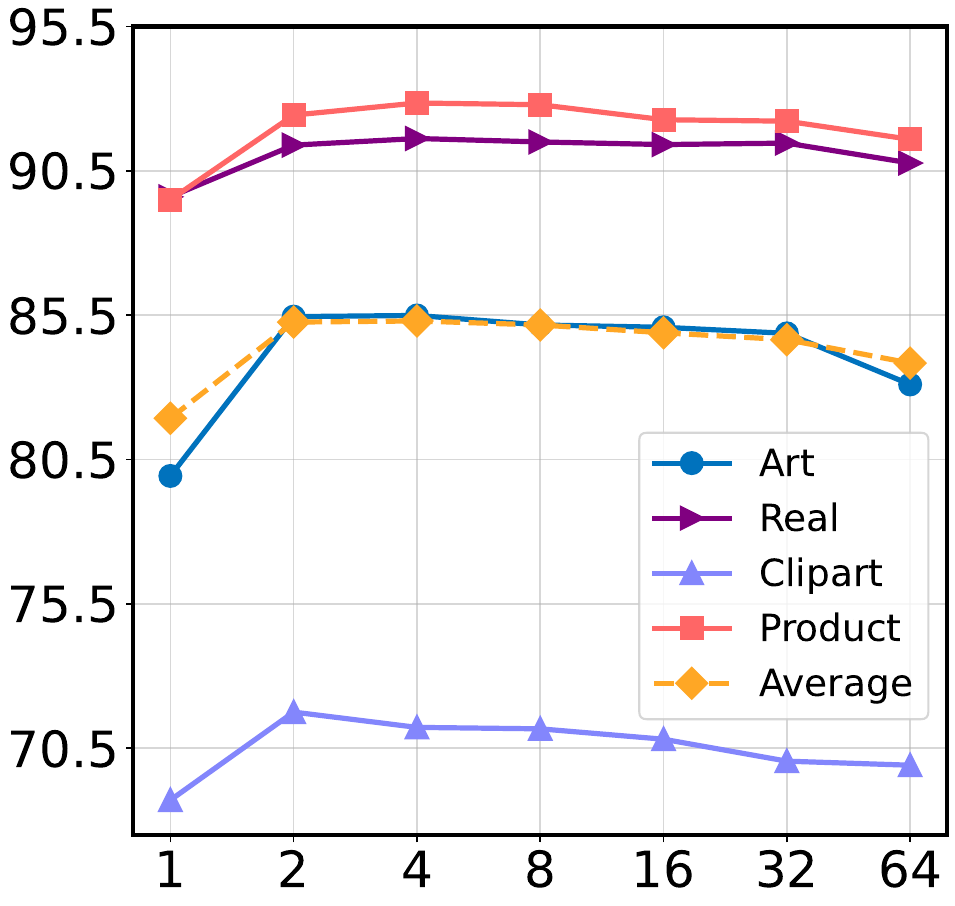}
		\label{p_nums}}
	\captionsetup{skip=10pt}
	\caption{Effect of the number of prompt tokens and experts.}
	\label{fig:prompt}
\end{figure}
Fig.~\ref{fig:prompt} illustrates the effect of varying the number of prompt tokens and experts on performance. 
In Fig.~\ref{p_length}, we fix the number of experts at 4 and incrementally increase prompt tokens from 1 to 64 to  examine their effect. 
As the number of prompt tokens increases, the average performance across four domains exhibits a corresponding improvement, achieving the best results at 32 tokens.
Beyond this point, a decline in performance is observed, likely due to overfitting to local client data distributions, which compromises generalization ability. 
We also observe that the ``Clipart" domain exhibits high sensitivity to this parameter, with minor adjustments causing substantial performance variations. 
In contrast, the other three domains follow performance trends that closely mirror the overall average.

To investigate the effect of the number of prompt experts, we fix the number of prompt tokens at 32 while varying the number of experts. 
As illustrated in Fig.~\ref{p_nums}, increasing the number of experts from 1 to 2 yields significant performance gains across all domains, with an average improvement of 3.33\%, from 81.93\% to 85.26\%.
Further increasing the number of experts beyond the optimal value of 4 leads to a gradual performance decline. This degradation likely arises because too many experts weaken individual specialization, resulting in redundancy and functional overlap among experts.
However, even with 64 experts, TRIP still outperforms the single-expert baseline by a significant margin of 1.93\%.
\subsection{Cost Analysis}
\begin{figure}[t]
	\centering
	\includegraphics[width=0.9\linewidth]{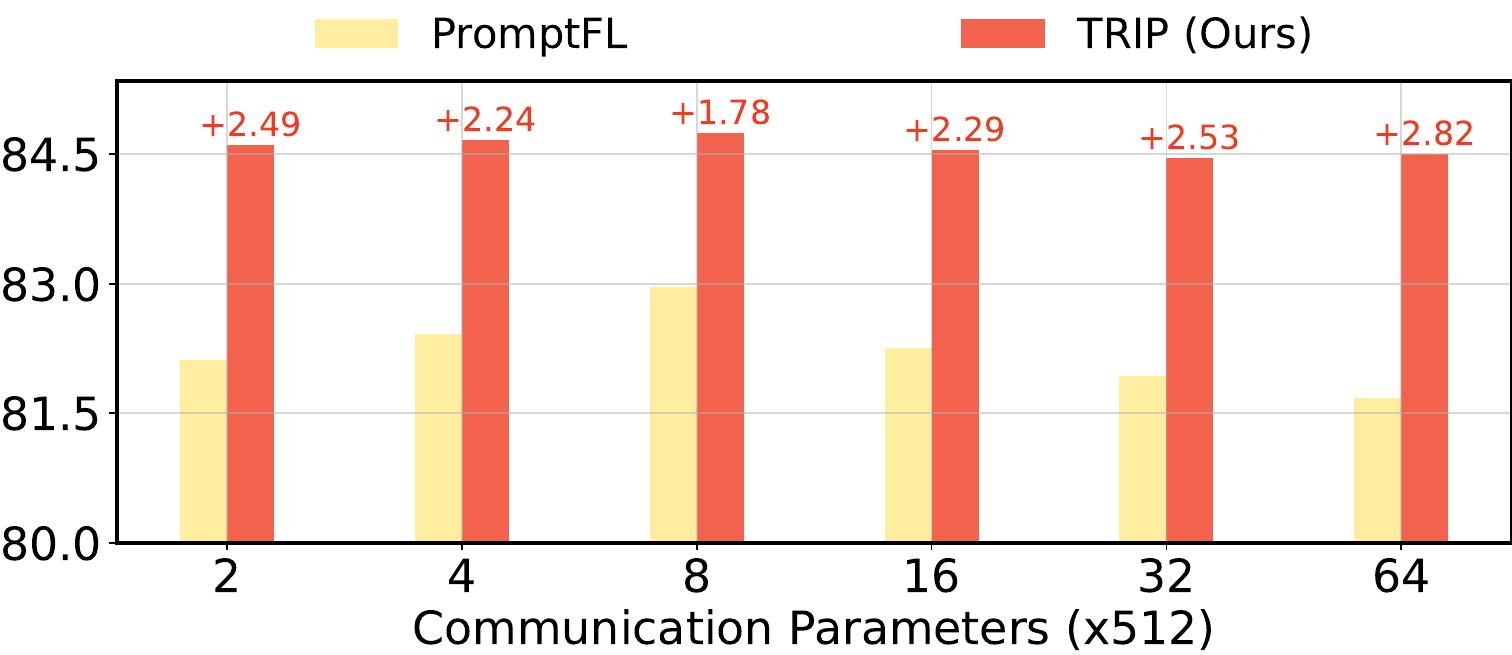}
	\captionsetup{skip=-1pt}
	\caption{Accuracy  of PromptFL and TRIP under equivalent communication parameters per round.}
	\label{fig:same_para_acc}
\end{figure}
As detailed in Subsection~\ref{subsec:performance}, our method achieves superior communication efficiency compared to existing methods.
Notably, PromptFL can be regarded as a special case of TRIP, where the number of experts and prompt tokens are set to 1 and 16, respectively. 
To highlight TRIP's advantages, we conduct a comparative performance evaluation under the same communication budgets.
For TRIP, we fix the number of prompt tokens at 1 and only vary  the number of experts. Conversely, for PromptFL, the number of prompt tokens is gradually increased.
As shown in Fig.~\ref{fig:same_para_acc}, TRIP consistently achieves accuracy improvements ranging from 1.78 to 2.82 percentage points across all tested communication parameter settings. 
Furthermore, PromptFL’s generalization performance degrades significantly as the number of parameters beyond 4,096, likely due to overfitting on local client data. 
In contrast, TRIP demonstrates better robustness to parameter scaling, confirming the effectiveness of its unbiased optimization strategy.

\begin{figure*}[t]
	\centering
	\includegraphics[width=0.95\linewidth]{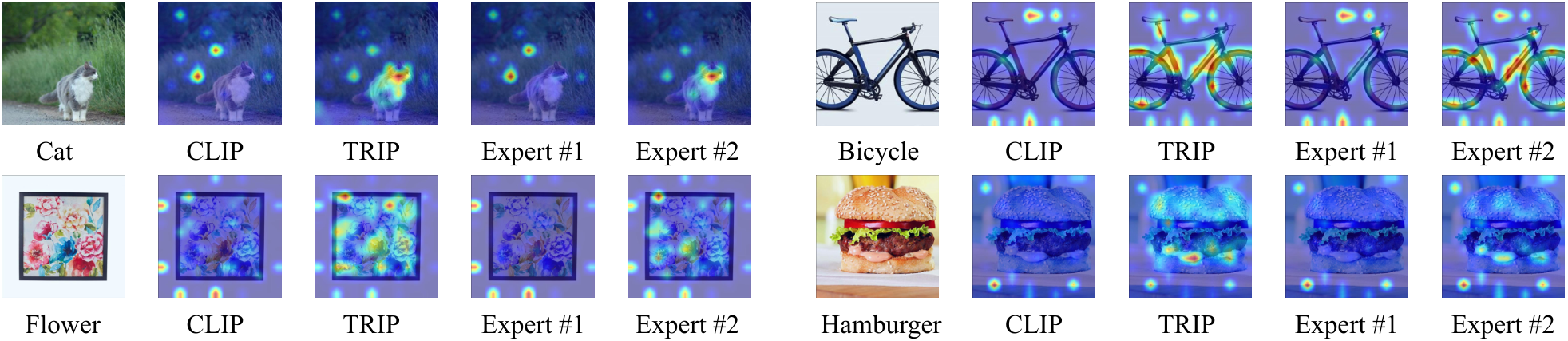}
	\captionsetup{skip=-1pt}
	\caption{The Grad-CAM visualization results for different prompting methods evaluated on  DomainNet.}
	\label{fig:vis}
\end{figure*}
\subsection{Visualization}
We employ Grad-CAM~\cite{selvaraju2017grad} to visualize how different prompting strategies influence the CLIP's attention toward discriminative image regions.
Our analysis compares three approaches: (1) CLIP’s fixed prompt, ``a photo of a [CLASS].", (2) TRIP’s mixture of prompts, and (3) individual prompt experts within TRIP.
As shown in Fig.~\ref{fig:vis}, TRIP’s mixture of prompts directs better focus towards objects compared to CLIP’s fixed prompt, which primarily activates background regions.
Furthermore, visualizations reveal that TRIP’s prompt experts exhibit distinct specializations. Specifically, Expert \#1 prioritizes contextual background information, whereas Expert \#2 focuses on semantically critical object areas.
\section{Conclusion}\label{sec:conclusions}
In this paper, we focus on federated domain generalization (FedDG), leveraging prompt learning and mixture of experts (MoE).
We propose using the MoE framework for instance-specific prompt synthesis, where multiple prompts are treated as distinct experts.
Specifically, we introduce a Token-level pRompt mIxture with Parameter-free routing framework for FedDG, called TRIP. Unlike existing federated MoE methods that assign entire images to experts, TRIP enables token-to-expert assignment, facilitating more effective capture of fine-grained regional image information. 
To ensure communication efficiency, TRIP incorporates a parameter-free routing mechanism based on a custom clustering algorithm and optimal transport. Additionally, TRIP introduces an unbiased learning objective to mitigate prompt experts' bias toward local data patterns. 
Extensive experiments demonstrate TRIP achieves new state-of-the-art results while communicating 1K parameters per round.
Despite its effectiveness, TRIP assumes that source and target clients share the same label space. In future work, we plan to extend our approach to open-set scenarios.
\bibliographystyle{IEEEtran}
\bibliography{Ref}

\begin{thebibliography}{10}
\providecommand{\url}[1]{#1}
\csname url@samestyle\endcsname
\providecommand{\newblock}{\relax}
\providecommand{\bibinfo}[2]{#2}
\providecommand{\BIBentrySTDinterwordspacing}{\spaceskip=0pt\relax}
\providecommand{\BIBentryALTinterwordstretchfactor}{4}
\providecommand{\BIBentryALTinterwordspacing}{\spaceskip=\fontdimen2\font plus
\BIBentryALTinterwordstretchfactor\fontdimen3\font minus
  \fontdimen4\font\relax}
\providecommand{\BIBforeignlanguage}[2]{{%
\expandafter\ifx\csname l@#1\endcsname\relax
\typeout{** WARNING: IEEEtran.bst: No hyphenation pattern has been}%
\typeout{** loaded for the language `#1'. Using the pattern for}%
\typeout{** the default language instead.}%
\else
\language=\csname l@#1\endcsname
\fi
#2}}
\providecommand{\BIBdecl}{\relax}
\BIBdecl

\bibitem{mcmahan2017communication}
B.~McMahan, E.~Moore, D.~Ramage, S.~Hampson, and B.~A. y~Arcas,
  ``Communication-efficient learning of deep networks from decentralized
  data,'' in \emph{Artificial intelligence and statistics}.\hskip 1em plus
  0.5em minus 0.4em\relax PMLR, 2017, pp. 1273--1282.

\bibitem{li2021survey}
Q.~Li, Z.~Wen, Z.~Wu, S.~Hu, N.~Wang, Y.~Li, X.~Liu, and B.~He, ``A survey on
  federated learning systems: Vision, hype and reality for data privacy and
  protection,'' \emph{IEEE Transactions on Knowledge and Data Engineering},
  vol.~35, no.~4, pp. 3347--3366, 2021.

\bibitem{zhou2024mixstyle}
K.~Zhou, Y.~Yang, Y.~Qiao, and T.~Xiang, ``Mixstyle neural networks for domain
  generalization and adaptation,'' \emph{International Journal of Computer
  Vision}, vol. 132, no.~3, pp. 822--836, 2024.

\bibitem{dayal2024madg}
A.~Dayal, V.~KB, L.~R. Cenkeramaddi, C.~Mohan, A.~Kumar, and
  V.~N~Balasubramanian, ``Madg: Margin-based adversarial learning for domain
  generalization,'' \emph{Advances in Neural Information Processing Systems},
  vol.~36, 2024.

\bibitem{liu2021feddg}
Q.~Liu, C.~Chen, J.~Qin, Q.~Dou, and P.-A. Heng, ``Feddg: Federated domain
  generalization on medical image segmentation via episodic learning in
  continuous frequency space,'' in \emph{Proceedings of the IEEE/CVF Conference
  on Computer Vision and Pattern Recognition}, 2021, pp. 1013--1023.

\bibitem{radford2021learning}
A.~Radford, J.~W. Kim, C.~Hallacy, A.~Ramesh, G.~Goh, S.~Agarwal, G.~Sastry,
  A.~Askell, P.~Mishkin, J.~Clark \emph{et~al.}, ``Learning transferable visual
  models from natural language supervision,'' in \emph{International conference
  on machine learning}.\hskip 1em plus 0.5em minus 0.4em\relax PMLR, 2021, pp.
  8748--8763.

\bibitem{li2022blip}
J.~Li, D.~Li, C.~Xiong, and S.~Hoi, ``Blip: Bootstrapping language-image
  pre-training for unified vision-language understanding and generation,'' in
  \emph{International conference on machine learning}.\hskip 1em plus 0.5em
  minus 0.4em\relax PMLR, 2022, pp. 12\,888--12\,900.

\bibitem{zhou2022learning}
K.~Zhou, J.~Yang, C.~C. Loy, and Z.~Liu, ``Learning to prompt for
  vision-language models,'' \emph{International Journal of Computer Vision},
  vol. 130, no.~9, pp. 2337--2348, 2022.

\bibitem{khattak2023maple}
M.~U. Khattak, H.~Rasheed, M.~Maaz, S.~Khan, and F.~S. Khan, ``Maple:
  Multi-modal prompt learning,'' in \emph{Proceedings of the IEEE/CVF
  Conference on Computer Vision and Pattern Recognition}, 2023, pp.
  19\,113--19\,122.

\bibitem{bai2024diprompt}
S.~Bai, J.~Zhang, S.~Li, S.~Guo, J.~Guo, J.~Hou, T.~Han, and X.~Lu, ``Diprompt:
  Disentangled prompt tuning for multiple latent domain generalization in
  federated learning,'' \emph{arXiv preprint arXiv:2403.08506}, 2024.

\bibitem{gong2024federated}
S.~Gong, C.~Cui, C.~Zhang, W.~Wang, X.~Nie, and L.~Zhu, ``Federated domain
  generalization via prompt learning and aggregation,'' \emph{arXiv preprint
  arXiv:2411.10063}, 2024.

\bibitem{guo2023promptfl}
T.~Guo, S.~Guo, J.~Wang, X.~Tang, and W.~Xu, ``Promptfl: Let federated
  participants cooperatively learn prompts instead of models-federated learning
  in age of foundation model,'' \emph{IEEE Transactions on Mobile Computing},
  2023.

\bibitem{luo2024mixture}
J.~Luo, C.~Chen, and S.~Wu, ``Mixture of experts made personalized: Federated
  prompt learning for vision-language models,'' \emph{arXiv preprint
  arXiv:2410.10114}, 2024.

\bibitem{feng2023towards}
C.-M. Feng, K.~Yu, N.~Liu, X.~Xu, S.~Khan, and W.~Zuo, ``Towards
  instance-adaptive inference for federated learning,'' in \emph{Proceedings of
  the IEEE/CVF International Conference on Computer Vision}, 2023, pp.
  23\,287--23\,296.

\bibitem{jacobs1991adaptive}
R.~A. Jacobs, M.~I. Jordan, S.~J. Nowlan, and G.~E. Hinton, ``Adaptive mixtures
  of local experts,'' \emph{Neural computation}, vol.~3, no.~1, pp. 79--87,
  1991.

\bibitem{dai2024deepseekmoe}
D.~Dai, C.~Deng, C.~Zhao, R.~Xu, H.~Gao, D.~Chen, J.~Li, W.~Zeng, X.~Yu, Y.~Wu
  \emph{et~al.}, ``Deepseekmoe: Towards ultimate expert specialization in
  mixture-of-experts language models,'' \emph{CoRR}, 2024.

\bibitem{wu2024mixture}
Z.~Wu, H.-Y. Huang, F.~Qu, and Y.~Wu, ``Mixture-of-prompt-experts for
  multi-modal semantic understanding,'' in \emph{Proceedings of the 2024 Joint
  International Conference on Computational Linguistics, Language Resources and
  Evaluation (LREC-COLING 2024)}, 2024, pp. 11\,381--11\,393.

\bibitem{monge1781memoire}
G.~Monge, ``M{\'e}moire sur la th{\'e}orie des d{\'e}blais et des remblais,''
  \emph{Mem. Math. Phys. Acad. Royale Sci.}, pp. 666--704, 1781.

\bibitem{khattak2023self}
M.~U. Khattak, S.~T. Wasim, M.~Naseer, S.~Khan, M.-H. Yang, and F.~S. Khan,
  ``Self-regulating prompts: Foundational model adaptation without
  forgetting,'' in \emph{Proceedings of the IEEE/CVF International Conference
  on Computer Vision}, 2023, pp. 15\,190--15\,200.

\bibitem{li2020federated}
T.~Li, A.~K. Sahu, M.~Zaheer, M.~Sanjabi, A.~Talwalkar, and V.~Smith,
  ``Federated optimization in heterogeneous networks,'' \emph{Proceedings of
  Machine learning and systems}, vol.~2, pp. 429--450, 2020.

\bibitem{malaviya2023fedfame}
S.~Malaviya, M.~Shukla, P.~Korat, and S.~Lodha, ``Fedfame: A data augmentation
  free framework based on model contrastive learning for federated
  semi-supervised learning,'' in \emph{Proceedings of the 38th ACM/SIGAPP
  Symposium on Applied Computing}, 2023, pp. 1114--1121.

\bibitem{karimireddy2020scaffold}
S.~P. Karimireddy, S.~Kale, M.~Mohri, S.~Reddi, S.~Stich, and A.~T. Suresh,
  ``Scaffold: Stochastic controlled averaging for federated learning,'' in
  \emph{International conference on machine learning}.\hskip 1em plus 0.5em
  minus 0.4em\relax PMLR, 2020, pp. 5132--5143.

\bibitem{wang2020federated}
H.~Wang, M.~Yurochkin, Y.~Sun, D.~Papailiopoulos, and Y.~Khazaeni, ``Federated
  learning with matched averaging,'' in \emph{International Conference on
  Learning Representations}, 2020.

\bibitem{wang2020tackling}
J.~Wang, Q.~Liu, H.~Liang, G.~Joshi, and H.~V. Poor, ``Tackling the objective
  inconsistency problem in heterogeneous federated optimization,''
  \emph{Advances in neural information processing systems}, vol.~33, pp.
  7611--7623, 2020.

\bibitem{li2021fedbn}
X.~Li, M.~Jiang, X.~Zhang, M.~Kamp, and Q.~Dou, ``Fedbn: Federated learning on
  non-iid features via local batch normalization,'' \emph{arXiv preprint
  arXiv:2102.07623}, 2021.

\bibitem{chen2024fair}
Y.~Chen, W.~Huang, and M.~Ye, ``Fair federated learning under domain skew with
  local consistency and domain diversity,'' in \emph{Proceedings of the
  IEEE/CVF Conference on Computer Vision and Pattern Recognition}, 2024, pp.
  12\,077--12\,086.

\bibitem{li2018domain}
H.~Li, S.~J. Pan, S.~Wang, and A.~C. Kot, ``Domain generalization with
  adversarial feature learning,'' in \emph{Proceedings of the IEEE conference
  on computer vision and pattern recognition}, 2018, pp. 5400--5409.

\bibitem{motiian2017unified}
S.~Motiian, M.~Piccirilli, D.~A. Adjeroh, and G.~Doretto, ``Unified deep
  supervised domain adaptation and generalization,'' in \emph{Proceedings of
  the IEEE international conference on computer vision}, 2017, pp. 5715--5725.

\bibitem{zhoudomain}
K.~Zhou, Y.~Yang, Y.~Qiao, and T.~Xiang, ``Domain generalization with
  mixstyle,'' in \emph{International Conference on Learning Representations}.

\bibitem{seo2020learning}
S.~Seo, Y.~Suh, D.~Kim, G.~Kim, J.~Han, and B.~Han, ``Learning to optimize
  domain specific normalization for domain generalization,'' in \emph{Computer
  Vision--ECCV 2020: 16th European Conference, Glasgow, UK, August 23--28,
  2020, Proceedings, Part XXII 16}.\hskip 1em plus 0.5em minus 0.4em\relax
  Springer, 2020, pp. 68--83.

\bibitem{li2018learning}
D.~Li, Y.~Yang, Y.-Z. Song, and T.~Hospedales, ``Learning to generalize:
  Meta-learning for domain generalization,'' in \emph{Proceedings of the AAAI
  conference on artificial intelligence}, vol.~32, no.~1, 2018.

\bibitem{dou2019domain}
Q.~Dou, D.~Coelho~de Castro, K.~Kamnitsas, and B.~Glocker, ``Domain
  generalization via model-agnostic learning of semantic features,''
  \emph{Advances in neural information processing systems}, vol.~32, 2019.

\bibitem{chen2023federated}
J.~Chen, M.~Jiang, Q.~Dou, and Q.~Chen, ``Federated domain generalization for
  image recognition via cross-client style transfer,'' in \emph{Proceedings of
  the IEEE/CVF Winter Conference on Applications of Computer Vision}, 2023, pp.
  361--370.

\bibitem{park2024stablefdg}
J.~Park, D.-J. Han, J.~Kim, S.~Wang, C.~Brinton, and J.~Moon, ``Stablefdg:
  style and attention based learning for federated domain generalization,''
  \emph{Advances in Neural Information Processing Systems}, vol.~36, 2024.

\bibitem{huang2023rethinking}
W.~Huang, M.~Ye, Z.~Shi, H.~Li, and B.~Du, ``Rethinking federated learning with
  domain shift: A prototype view,'' in \emph{2023 IEEE/CVF Conference on
  Computer Vision and Pattern Recognition (CVPR)}.\hskip 1em plus 0.5em minus
  0.4em\relax IEEE, 2023, pp. 16\,312--16\,322.

\bibitem{zhang2023federated}
R.~Zhang, Q.~Xu, J.~Yao, Y.~Zhang, Q.~Tian, and Y.~Wang, ``Federated domain
  generalization with generalization adjustment,'' in \emph{Proceedings of the
  IEEE/CVF Conference on Computer Vision and Pattern Recognition}, 2023, pp.
  3954--3963.

\bibitem{zhou2022conditional}
K.~Zhou, J.~Yang, C.~C. Loy, and Z.~Liu, ``Conditional prompt learning for
  vision-language models,'' in \emph{Proceedings of the IEEE/CVF conference on
  computer vision and pattern recognition}, 2022, pp. 16\,816--16\,825.

\bibitem{chenplot}
G.~Chen, W.~Yao, X.~Song, X.~Li, Y.~Rao, and K.~Zhang, ``Plot: Prompt learning
  with optimal transport for vision-language models,'' in \emph{The Eleventh
  International Conference on Learning Representations}.

\bibitem{wang2024tuning}
D.~Wang, M.~Li, X.~Liu, M.~Xu, B.~Chen, and H.~Zhang, ``Tuning multi-mode
  token-level prompt alignment across modalities,'' \emph{Advances in Neural
  Information Processing Systems}, vol.~36, 2024.

\bibitem{li2024global}
H.~Li, W.~Huang, J.~Wang, and Y.~Shi, ``Global and local prompts cooperation
  via optimal transport for federated learning,'' in \emph{Proceedings of the
  IEEE/CVF Conference on Computer Vision and Pattern Recognition}, 2024, pp.
  12\,151--12\,161.

\bibitem{fedus2022switch}
W.~Fedus, B.~Zoph, and N.~Shazeer, ``Switch transformers: Scaling to trillion
  parameter models with simple and efficient sparsity,'' \emph{Journal of
  Machine Learning Research}, vol.~23, no. 120, pp. 1--39, 2022.

\bibitem{wumixture}
X.~Wu, S.~Huang, and F.~Wei, ``Mixture of lora experts,'' in \emph{The Twelfth
  International Conference on Learning Representations}.

\bibitem{shahbaba2009nonlinear}
B.~Shahbaba and R.~Neal, ``Nonlinear models using dirichlet process mixtures.''
  \emph{Journal of Machine Learning Research}, vol.~10, no.~8, 2009.

\bibitem{shazeer2017outrageously}
N.~Shazeer, A.~Mirhoseini, K.~Maziarz, A.~Davis, Q.~Le, G.~Hinton, and J.~Dean,
  ``Outrageously large neural networks: The sparsely-gated mixture-of-experts
  layer,'' \emph{arXiv preprint arXiv:1701.06538}, 2017.

\bibitem{dun2023fedjets}
C.~Dun, M.~H. Garcia, G.~Zheng, A.~Awadallah, R.~Sim, A.~Kyrillidis, and
  D.~Dimitriadis, ``Fedjets: Efficient just-in-time personalization with
  federated mixture of experts,'' in \emph{R0-FoMo: Robustness of Few-shot and
  Zero-shot Learning in Large Foundation Models}, 2023.

\bibitem{reisser2021federated}
M.~Reisser, C.~Louizos, E.~Gavves, and M.~Welling, ``Federated mixture of
  experts,'' \emph{arXiv preprint arXiv:2107.06724}, 2021.

\bibitem{radford2019language}
A.~Radford, J.~Wu, R.~Child, D.~Luan, D.~Amodei, I.~Sutskever \emph{et~al.},
  ``Language models are unsupervised multitask learners,'' \emph{OpenAI blog},
  vol.~1, no.~8, p.~9, 2019.

\bibitem{kuhn1955hungarian}
H.~W. Kuhn, ``The hungarian method for the assignment problem,'' \emph{Naval
  research logistics quarterly}, vol.~2, no. 1-2, pp. 83--97, 1955.

\bibitem{li2017deeper}
D.~Li, Y.~Yang, Y.-Z. Song, and T.~M. Hospedales, ``Deeper, broader and artier
  domain generalization,'' in \emph{Proceedings of the IEEE international
  conference on computer vision}, 2017, pp. 5542--5550.

\bibitem{venkateswara2017deep}
H.~Venkateswara, J.~Eusebio, S.~Chakraborty, and S.~Panchanathan, ``Deep
  hashing network for unsupervised domain adaptation,'' in \emph{Proceedings of
  the IEEE conference on computer vision and pattern recognition}, 2017, pp.
  5018--5027.

\bibitem{fang2013unbiased}
C.~Fang, Y.~Xu, and D.~N. Rockmore, ``Unbiased metric learning: On the
  utilization of multiple datasets and web images for softening bias,'' in
  \emph{Proceedings of the IEEE International Conference on Computer Vision},
  2013, pp. 1657--1664.

\bibitem{peng2019moment}
X.~Peng, Q.~Bai, X.~Xia, Z.~Huang, K.~Saenko, and B.~Wang, ``Moment matching
  for multi-source domain adaptation,'' in \emph{Proceedings of the IEEE/CVF
  international conference on computer vision}, 2019, pp. 1406--1415.

\bibitem{huang2020self}
Z.~Huang, H.~Wang, E.~P. Xing, and D.~Huang, ``Self-challenging improves
  cross-domain generalization,'' in \emph{Computer vision--ECCV 2020: 16th
  European conference, Glasgow, UK, August 23--28, 2020, proceedings, part II
  16}.\hskip 1em plus 0.5em minus 0.4em\relax Springer, 2020, pp. 124--140.

\bibitem{xu2021fourier}
Q.~Xu, R.~Zhang, Y.~Zhang, Y.~Wang, and Q.~Tian, ``A fourier-based framework
  for domain generalization,'' in \emph{Proceedings of the IEEE/CVF conference
  on computer vision and pattern recognition}, 2021, pp. 14\,383--14\,392.

\bibitem{cha2021swad}
J.~Cha, S.~Chun, K.~Lee, H.-C. Cho, S.~Park, Y.~Lee, and S.~Park, ``Swad:
  Domain generalization by seeking flat minima,'' \emph{Advances in Neural
  Information Processing Systems}, vol.~34, pp. 22\,405--22\,418, 2021.

\bibitem{zheng2022prompt}
Z.~Zheng, X.~Yue, K.~Wang, and Y.~You, ``Prompt vision transformer for domain
  generalization,'' \emph{arXiv preprint arXiv:2208.08914}, 2022.

\bibitem{zhou2024hcvp}
G.~Zhou, Z.~Han, S.~Chen, B.~Huang, L.~Zhu, T.~Liu, L.~Yao, and K.~Zhang,
  ``Hcvp: Leveraging hierarchical contrastive visual prompt for domain
  generalization,'' \emph{IEEE Transactions on Multimedia}, 2024.

\bibitem{nguyen2022fedsr}
A.~T. Nguyen, P.~Torr, and S.~N. Lim, ``Fedsr: A simple and effective domain
  generalization method for federated learning,'' \emph{Advances in Neural
  Information Processing Systems}, vol.~35, pp. 38\,831--38\,843, 2022.

\bibitem{zhang2021federated}
L.~Zhang, X.~Lei, Y.~Shi, H.~Huang, and C.~Chen, ``Federated learning with
  domain generalization,'' \emph{arXiv preprint arXiv:2111.10487}, 2021.

\bibitem{lu2023fedclip}
W.~Lu, H.~Xixu, J.~Wang, and X.~Xie, ``Fedclip: Fast generalization and
  personalization for clip in federated learning,'' in \emph{ICLR 2023 Workshop
  on Trustworthy and Reliable Large-Scale Machine Learning Models}, 2023.

\bibitem{su2024federated}
S.~Su, M.~Yang, B.~Li, and X.~Xue, ``Federated adaptive prompt tuning for
  multi-domain collaborative learning,'' in \emph{Proceedings of the AAAI
  Conference on Artificial Intelligence}, vol.~38, no.~13, 2024, pp.
  15\,117--15\,125.

\bibitem{dosovitskiy2020image}
A.~Dosovitskiy, L.~Beyer, A.~Kolesnikov, D.~Weissenborn, X.~Zhai,
  T.~Unterthiner, M.~Dehghani, M.~Minderer, G.~Heigold, S.~Gelly \emph{et~al.},
  ``An image is worth 16x16 words: Transformers for image recognition at
  scale,'' in \emph{International Conference on Learning Representations},
  2020.

\bibitem{selvaraju2017grad}
R.~R. Selvaraju, M.~Cogswell, A.~Das, R.~Vedantam, D.~Parikh, and D.~Batra,
  ``Grad-cam: Visual explanations from deep networks via gradient-based
  localization,'' in \emph{Proceedings of the IEEE international conference on
  computer vision}, 2017, pp. 618--626.

\end{thebibliography}
\begin{IEEEbiography}[{\includegraphics[width=1in,height=1.25in,clip,keepaspectratio]{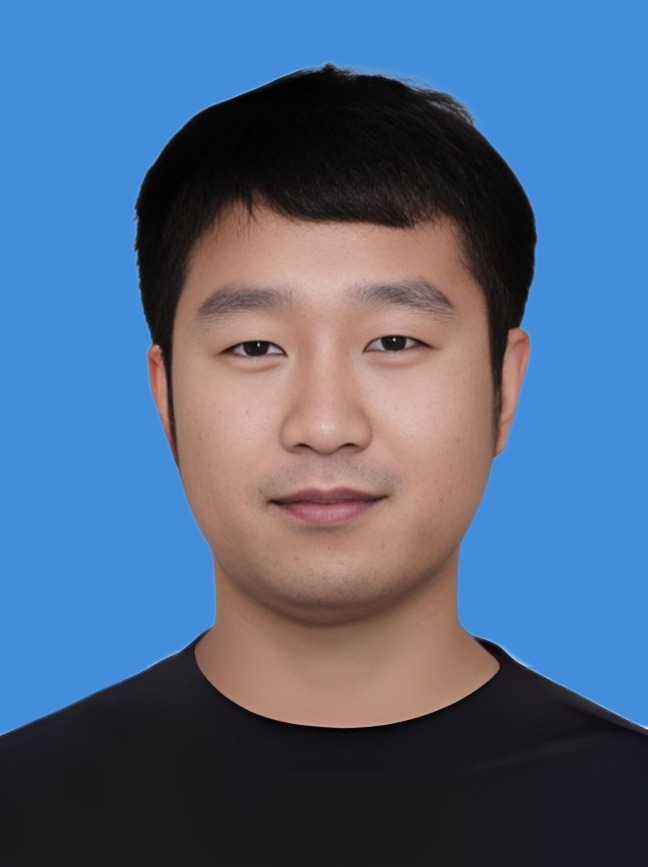}}]{Shuai Gong} received his master degree in computer
	science and technology from Shandong Jiaotong
	University, China, in 2023. During this time, he
	conducted research on natural language processing,
	focusing on automatic summarization. He is currently
	pursuing a Ph.D. in statistics at Shandong University
	of Finance and Economic, China. His research
	interests include transfer learning and federated
	learning.
\end{IEEEbiography}
\begin{IEEEbiography}[{\includegraphics[width=1in,height=1.25in,clip,keepaspectratio]{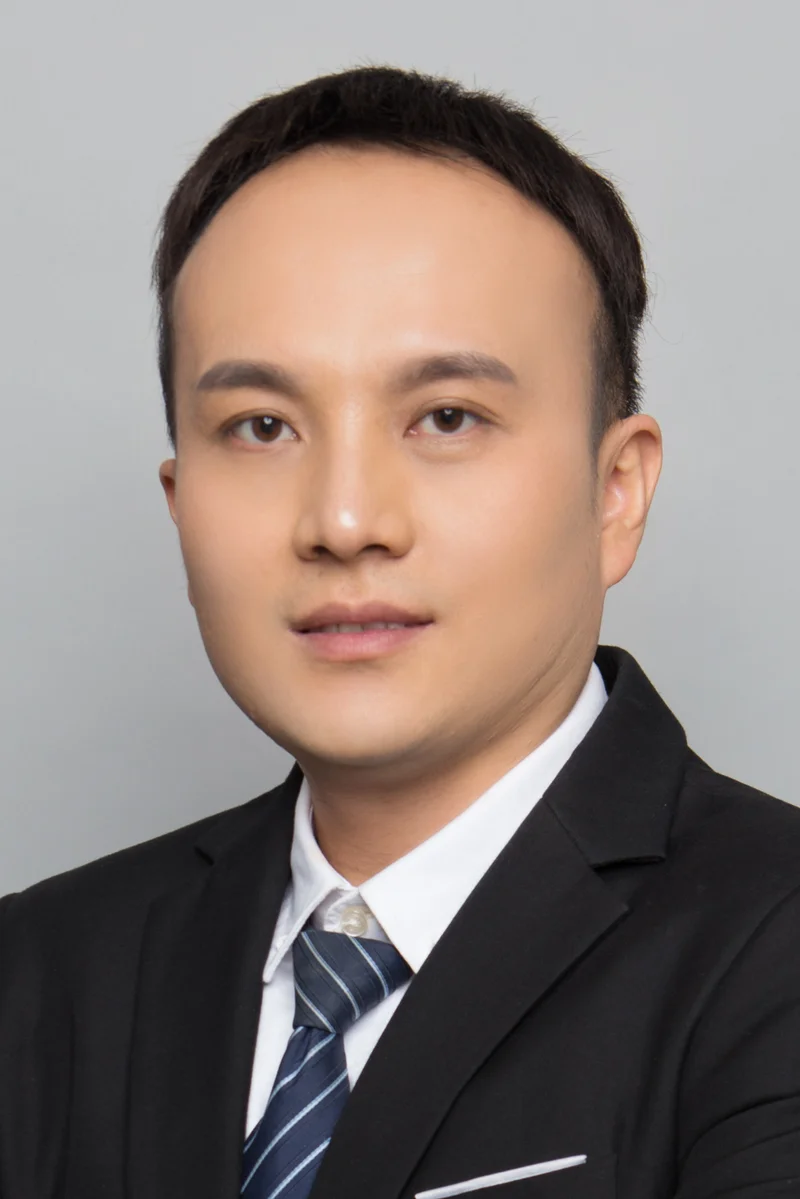}}]{Chaoran Cui} received his Ph.D. degree in 2015 in computer science from Shandong University, China. Prior to that, he received his B.E. degree
	in software engineering from Shandong University in 2010. During 2015-2016, he was a research fellow at Singapore Management University. He is now a professor with School of Computer Science and Technology, Shandong University of Finance and
	Economics, China. His research interests include machine learning, data mining, and multimedia processing.
\end{IEEEbiography}
\begin{IEEEbiography}[{\includegraphics[width=1in,height=1.25in,clip,keepaspectratio]{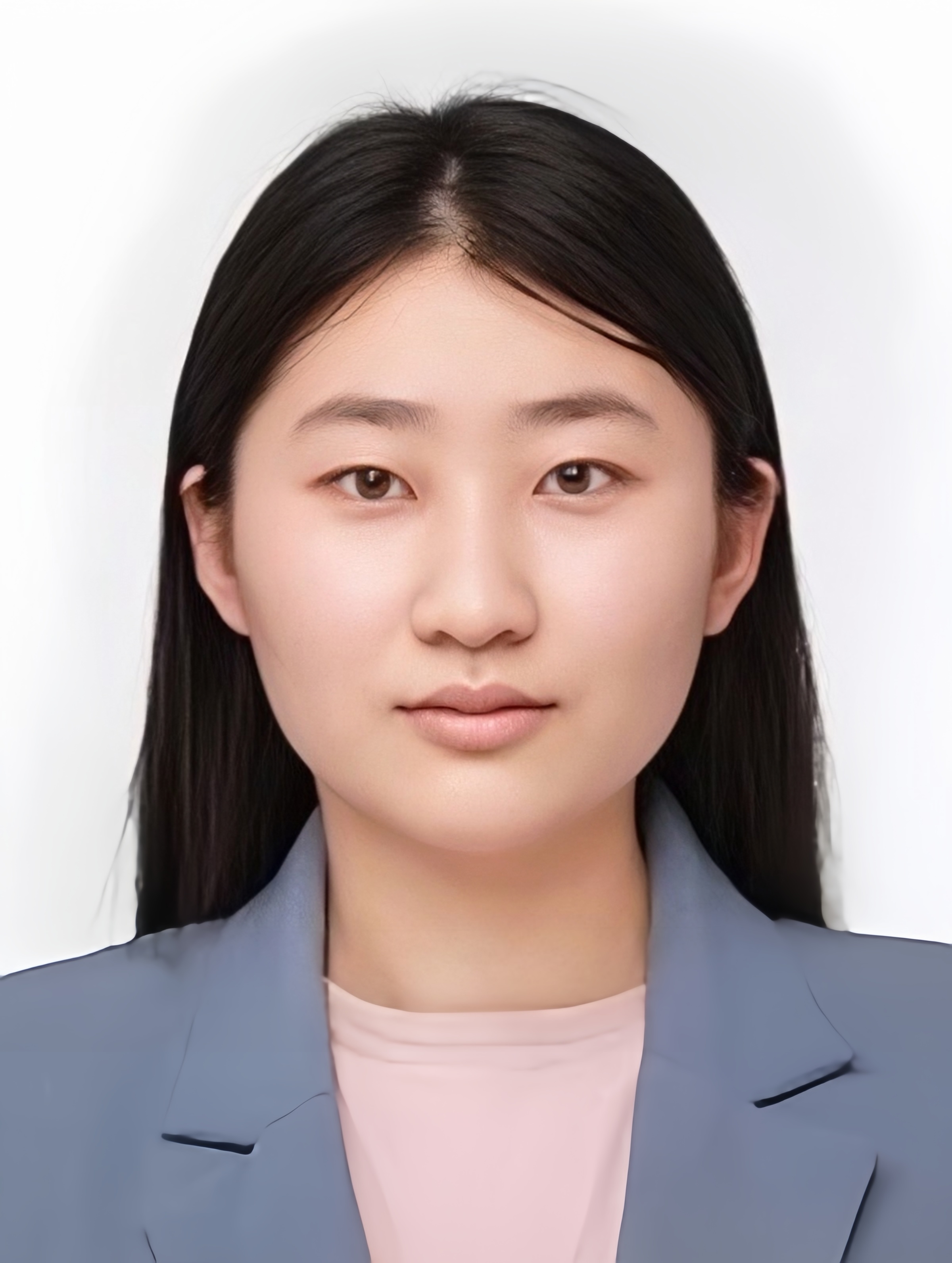}}]{Xiaolin Dong} is currently pursuing a Ph.D. in statistics at Shandong University of Finance and Economic, China.  Her research interests include prompt learning, black-box optimization, and transfer learning.
\end{IEEEbiography}
\begin{IEEEbiography}[{\includegraphics[width=1in,height=1.25in,clip,keepaspectratio]{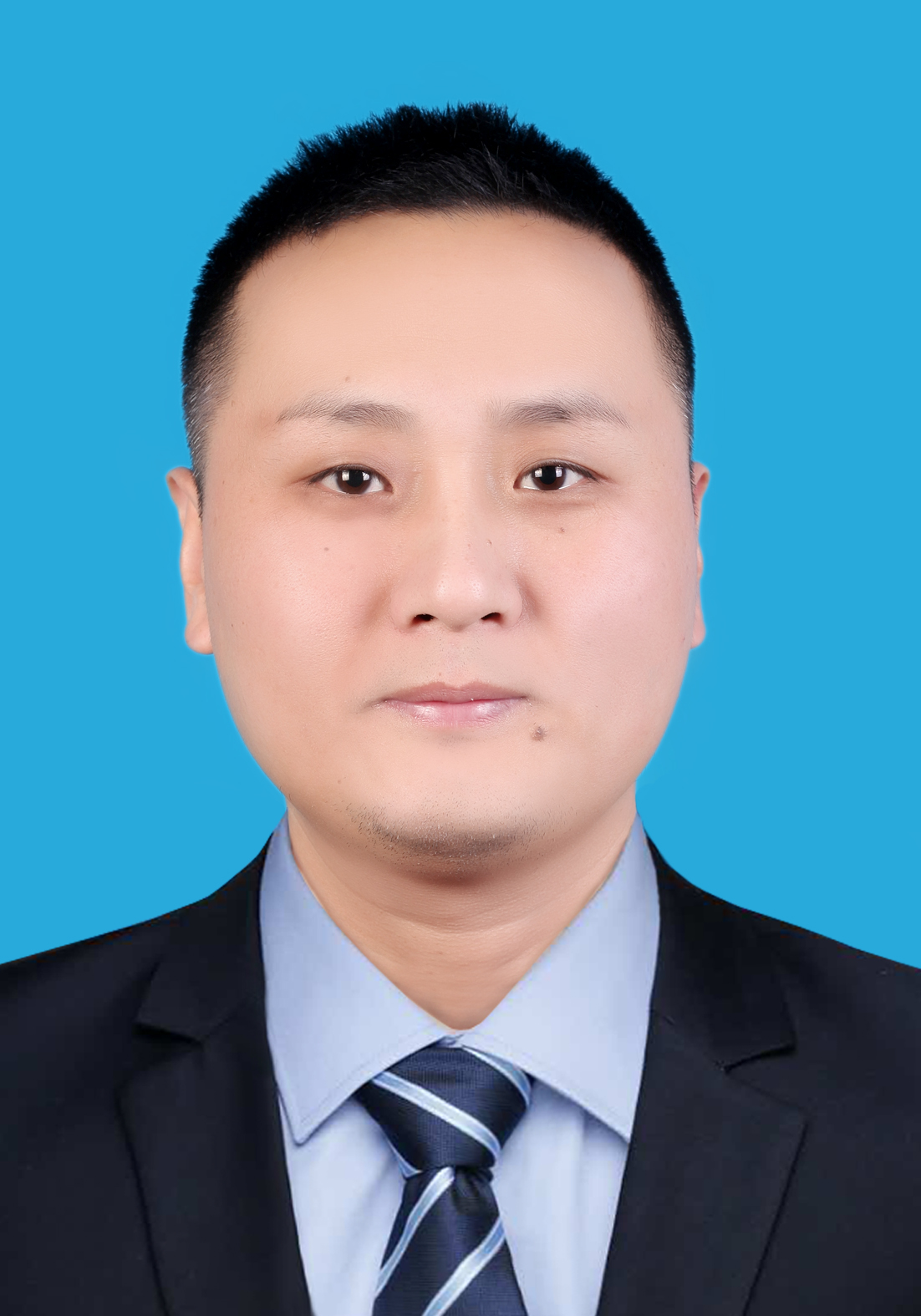}}]{Xiushan Nie} (Senior Member, IEEE) received the Ph.D. degree from Shandong University in 2011.
He was a Research Fellow with the University of Missouri, Columbia, MO, USA, from 2013 to 2014, under the supervision of Prof. Wenjun (Kevin) Zeng. He is currently a Full Professor with the School of Computer Science and Technology, Shandong Jianzhu University, China. His research interests include spatio-temporal intelligence, machine learning and computer vision.
\end{IEEEbiography}
\begin{IEEEbiography}[{\includegraphics[width=1in,height=1.25in,clip,keepaspectratio]{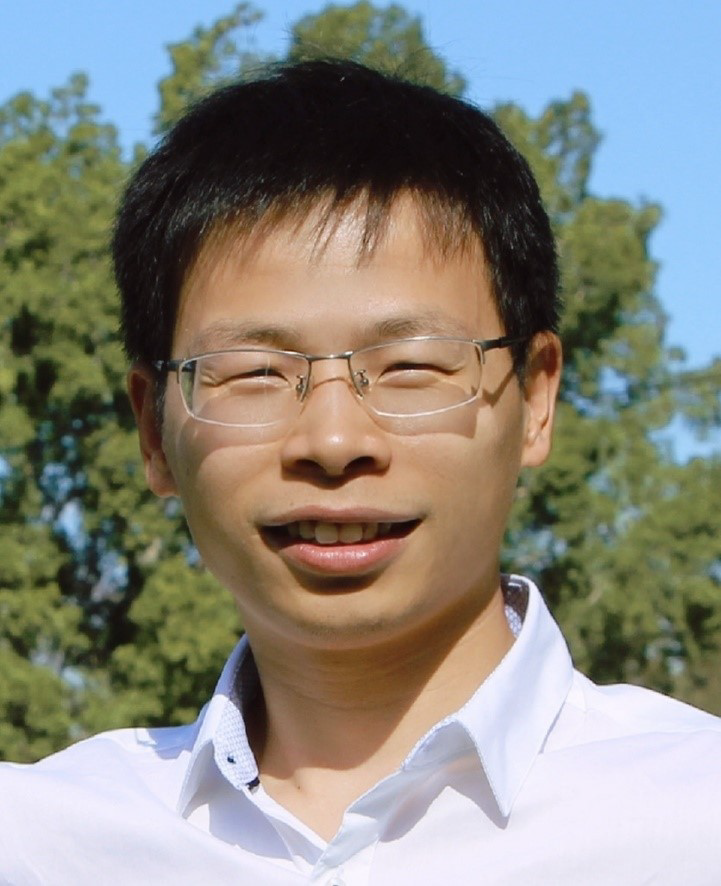}}]{Lei Zhu} (Senior Member, IEEE) received his B.Eng. and Ph.D. degrees
	from Wuhan University of Technology in 2009 and
	Huazhong University Science and Technology in
	2015, respectively. He was a Research Fellow at
	the University of Queensland (2016-2017). He is
	currently a professor with the College of Electronics
	and Information Engineering, Tongji University,
	Shanghai, China. His research interests are in the
	area of large-scale multimedia content analysis and
	retrieval.
\end{IEEEbiography}
\begin{IEEEbiography}[{\includegraphics[width=1in,height=1.25in,clip,keepaspectratio]{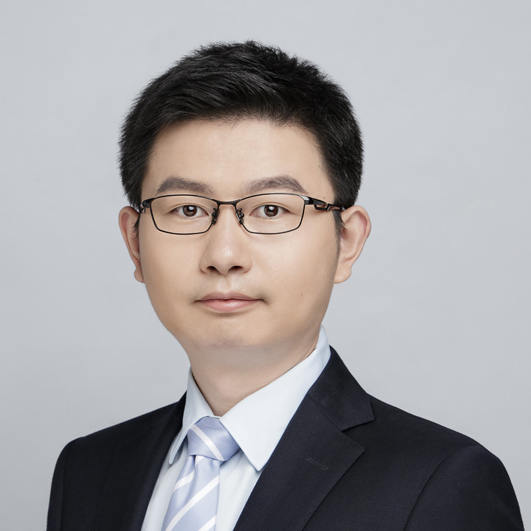}}]{Xiaojun Chang} (Senior Member, IEEE) is a Professor with the Department of Electronic Engineering and Information Science, University of Science and Technology of China (USTC). He is also a visiting Professor at Department of Computer Vision, Mohamed bin Zayed University of Artificial Intelligence (MBZUAI). He was an ARC Discovery Early Career Researcher Award (DECRA) Fellow between 2019-2021. After graduation, he was worked as a Postdoc Research Associate in School of Computer Science, Carnegie Mellon University, a Senior Lecturer in Faculty of Information Technology, Monash University, and an Associate Professor in School of Computing Technologies, RMIT University. He mainly worked on exploring multiple signals for automatic content analysis in unconstrained or surveillance videos and has achieved top performance in various international competitions. He received his Ph.D. degree from University of Technology Sydney. His research focus in this period was mainly on developing machine learning algorithms and applying them to multimedia analysis and computer vision.
\end{IEEEbiography}
\vfill
\end{document}